\documentclass[10pt,twocolumn,letterpaper]{article}
\usepackage[accsupp]{axessibility} 
\usepackage{cvpr}              
\definecolor{cvprblue}{rgb}{0.21,0.49,0.74}
\usepackage[pagebackref,breaklinks,colorlinks,allcolors=cvprblue]{hyperref}
\usepackage[T1]{fontenc}
\usepackage{graphicx}
\usepackage{multirow}
\usepackage{booktabs}
\usepackage{xcolor}
\usepackage{colortbl}
\usepackage{microtype}      
\usepackage{subcaption}
\usepackage{caption}
\usepackage{amsfonts}       
\usepackage{nicefrac}       
\usepackage{float} 
\usepackage{algorithm}
\usepackage{adjustbox}
\usepackage{pifont}
\usepackage{amsmath}
\usepackage{tikz}
\usepackage{listings}
\usetikzlibrary{positioning}
\usepackage{cancel}
\usetikzlibrary{calc}
\usetikzlibrary{arrows.meta}
\usepackage[most]{tcolorbox}
\usepackage[normalem]{ulem} 
\usepackage{algorithm, algpseudocode}
\usepackage{url}
\usepackage{color}
\usepackage{makecell}
\usepackage{cuted} 
\usepackage{tikz}
\usetikzlibrary{calc}
\usepackage{tikz}
\usetikzlibrary{positioning}

\def\paperID{33827} 
\def\confName{CVPR}
\def\confYear{2026}

\title{Finding Distributed Object-Centric Properties in Self-Supervised Transformers}

\author{
        Samyak Rawlekar$^{1}$ \quad 
        Amitabh Swain$^{1}$ \quad
        Yujun Cai$^{2}$ \quad 
        Yiwei Wang$^{3}$ \\ 
        Ming-Hsuan Yang$^{3}$ \quad 
        Narendra Ahuja$^{1}$
     \\ \\
        $^{1}$University of Illinois Urbana-Champaign \enskip 
        $^{2}$University of Queensland \enskip 
        $^{3}$UC Merced 
}

\begin{document}
\maketitle

\begin{strip}
    \centering
    \resizebox{\textwidth}{!}{%
    \begin{tabular}{@{\hskip 0pt}c@{\hskip 2pt}c@{\hskip 2pt}c@{\hskip 2pt}c@{\hskip 2pt}c@{\hskip 0pt}}
    \textit{Input Image} & \textit{Query Similarity} ($A_q$)  & \textit{Key Similarity}  ($A_k$)& \textit{Value Similarity} ($A_v$) & \textcolor{purple}{\textit{Ensemble Similarity}} \\
        \includegraphics[width=0.20\textwidth]{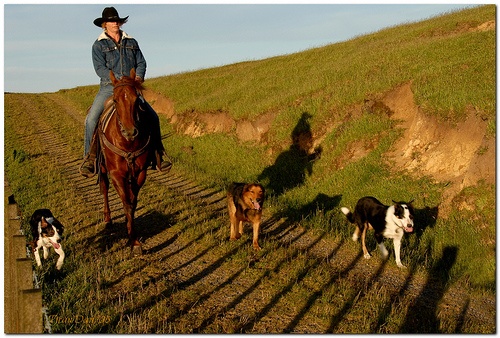} &
        \includegraphics[width=0.20\textwidth]{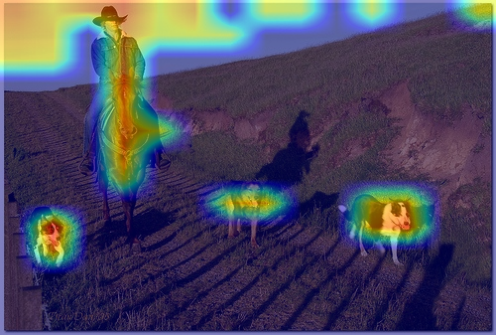} &
        \includegraphics[width=0.20\textwidth]{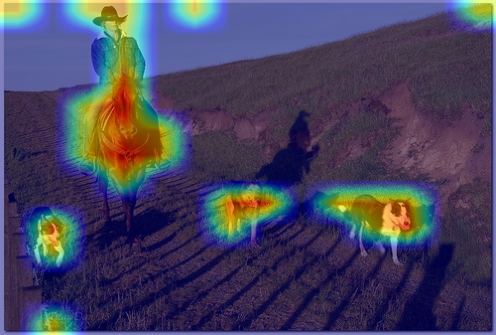} &
        \includegraphics[width=0.20\textwidth]{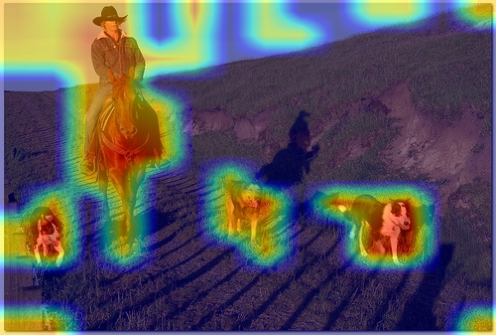} &
        \includegraphics[width=0.20\textwidth]{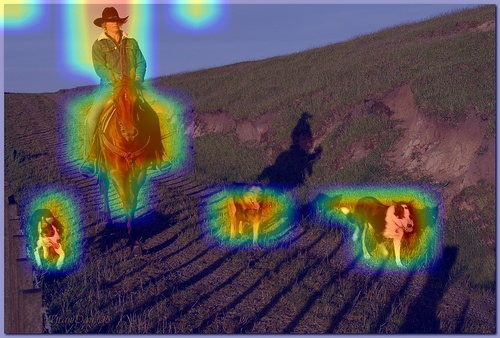} \\
        \includegraphics[width=0.20\textwidth]{figs/teaser/heatmaps_fig4/COCO_val2014_000000139605.jpg} &
        \includegraphics[width=0.20\textwidth]{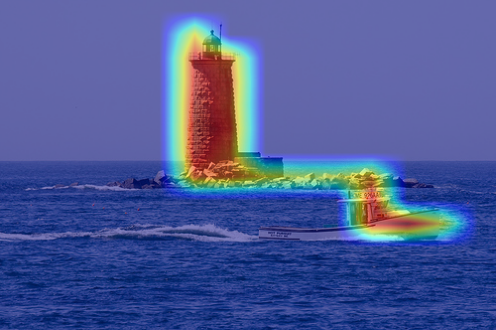} &
        \includegraphics[width=0.20\textwidth]{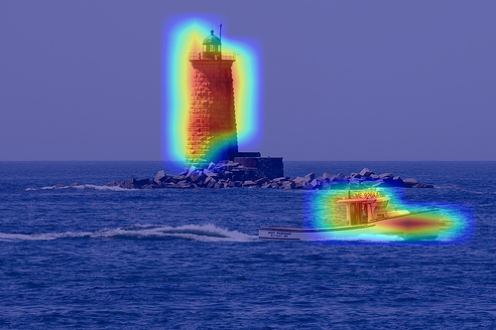} &
        \includegraphics[width=0.20\textwidth]{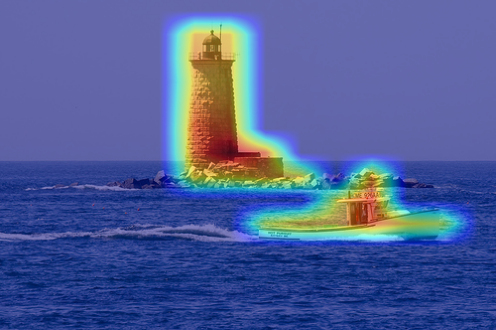} &
        \includegraphics[width=0.20\textwidth]{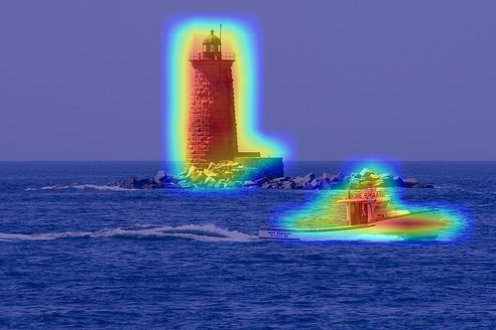} \\
        
    \end{tabular}
    }
    \captionof{figure}{\textbf{Object-centric information is encoded in patch-level interactions.} We visualize the inter-patch similarity maps ($A_q, A_k, A_v$) computed from the Query ($Q$), Key ($K$), and Value ($V$) representations of patch tokens. Each component captures a complementary view of object structure: Query Similarity reveals which patches seek similar information, Key Similarity shows which patches offer similar context, and Value Similarity identifies patches with similar content. The Ensemble aggregates all three components, producing foreground-background separation and precise object localization. For visualization, we invert the similarity maps so objects appear bright.}
    \label{fig:teaser}
\end{strip}

\begin{abstract}
Self-supervised Vision Transformers (ViTs) like DINO show an emergent ability to discover objects, typically observed in \texttt{[CLS]} token attention maps of the final layer. However, these maps often contain spurious activations resulting in poor localization of objects. This is because the \texttt{[CLS]} token, trained on an image-level objective, summarizes the entire image instead of focusing on objects. This aggregation dilutes the object-centric information existing in the local, patch-level interactions. We analyze this by computing inter-patch similarity using patch-level attention components (query, key, and value) across all layers. We find that: (1) Object-centric properties are encoded in the similarity maps derived from all three components ($q, k, v$), unlike prior work that uses only key features or the \texttt{[CLS]} token. (2) This object-centric information is distributed across the network, not just confined to the final layer. Based on these insights, we introduce Object-DINO, a training-free method that extracts this distributed object-centric information. Object-DINO clusters attention heads across all layers based on the similarities of their patches and automatically identifies the object-centric cluster corresponding to all objects. We demonstrate Object-DINO's effectiveness on two applications: enhancing unsupervised object discovery (+3.6 to +12.4 CorLoc gains) and mitigating object hallucination in Multimodal Large Language Models by providing visual grounding. Our results demonstrate that using this distributed object-centric information improves downstream tasks without additional training.
\end{abstract}


\section{Introduction}
\label{sec:intro}
Self-supervised Vision Transformers (ViTs), such as DINO \cite{dino}, have demonstrated an emergent ability to discover objects without any explicit supervision. This property is observed when the \texttt{[CLS]} token is used as a query in the self-attention mechanism of the final layer, producing attention maps that highlight salient object regions. However, DINO's training objective is global, image-level, and not a local-level feature matching. Consequently, the \texttt{[CLS]} token must learn to encode holistic image cues (e.g., texture, edges, and context) rather than focus on precise object regions. This creates a tension between the global and the desired local objectives, resulting in attention maps with poor localization/omission of objects, limiting their reliability for object discovery.

This limitation motivates us to reconsider where object-centric information truly resides in self-supervised ViTs. The \texttt{[CLS]} token forms its global summary by aggregating patch tokens through the self-attention mechanism. However, for this summary to capture semantic richness (as self-supervised learning objective requires), the self-attention mechanism must establish correspondences between individual patches based on their visual similarity and context. This means patches belonging to the same object naturally learn to attend to each other, forming coherent clusters that reflect object structure. This suggests that object-centric information may be more directly accessible via the local, patch-level groupings. This leads us to ask: \textbf{How to use the patch-level interactions to localize objects?}

To answer this question, we ignore the \texttt{[CLS]} token and compute inter-patch similarity directly from the patch-level attention components: query ($q$), key ($k$), and value ($v$). For each component type $r \in \{q, k, v\}$, we construct a normalized similarity matrix $A_r^{\ell,h}$ (Eq.~\ref{eq:self_sim}) in terms of dot products followed by softmax. Each matrix captures a distinct aspect of patch relationships: $A_q$ identifies which patches are looking for similar things, $A_k$ reveals patches offering similar contextual cues, and $A_v$ highlights patches containing similar content. As shown in Fig. \ref{fig:teaser}, and as we will see in Sec. \ref{sec:observation1}, these maps demonstrate strong localization properties.

Our finding that $A_q, A_k,$ and $A_v$ all contain strong localization properties reveals that prior work, such as TokenCut \cite{tokencut}, leveraged only a part of the available object-centric information by focusing solely on the key ($k$) features of the final layer. This suggests that we could obtain richer information by combining the three distinct components. However, it remains unclear how to produce this unified signal from them. Furthermore, prior work~\cite{dino,simeoni2021localizing,tokencut} has 
established that the final layer exhibits the strongest object-centric 
behavior. This raises a follow-up question: \textbf{Does the localization property emerge only at the final layer?}  If object representations emerge hierarchically through the network structure, then object-centric heads may be distributed across multiple layers. Moreover, even within the final layer, different heads may specialize in different functions, some capturing object-centric information while others encode different visual cues. If true, naively aggregating all heads would introduce noise from non-object-centric heads, while ignoring intermediate layers would discard valuable object-centric information.

\begin{figure*}[tp]
    \centering
    \resizebox{\textwidth}{!}{%
        \begin{tabular}{ccc}
            \begin{minipage}{0.32\textwidth}
                \centering
                \includegraphics[width=\linewidth,height=0.75\linewidth,keepaspectratio]{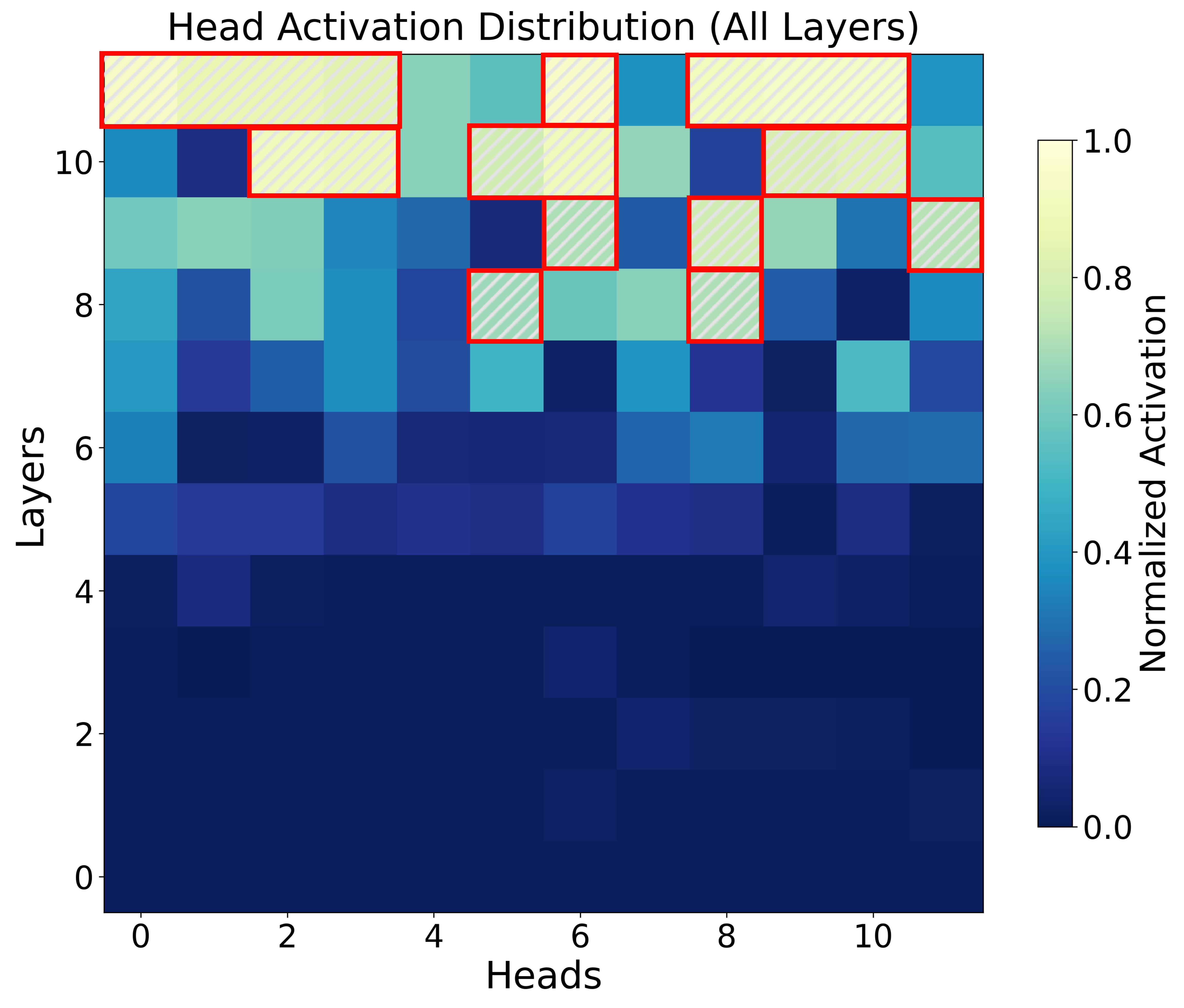}
                \caption*{(a) Head contribution for complete network}
            \end{minipage} &
            \begin{minipage}{0.33\textwidth}
                \centering
                \includegraphics[width=\linewidth,height=0.85\linewidth,keepaspectratio]{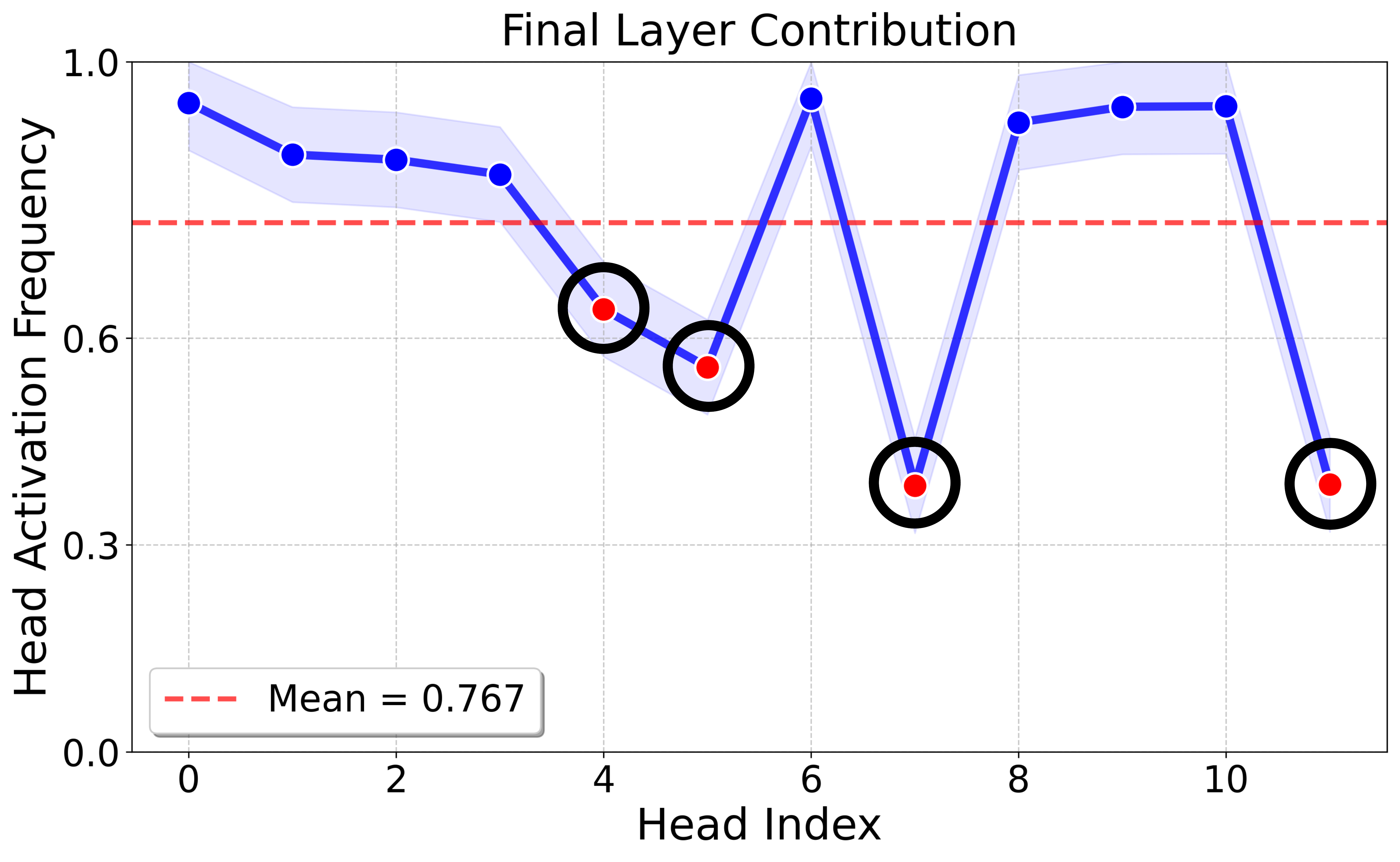}
                \caption*{(b) Heads contribution from final layer.}
            \end{minipage} &
            \begin{minipage}{0.33\textwidth}
                \centering
                \includegraphics[width=0.8\linewidth,height=0.8\linewidth,keepaspectratio]{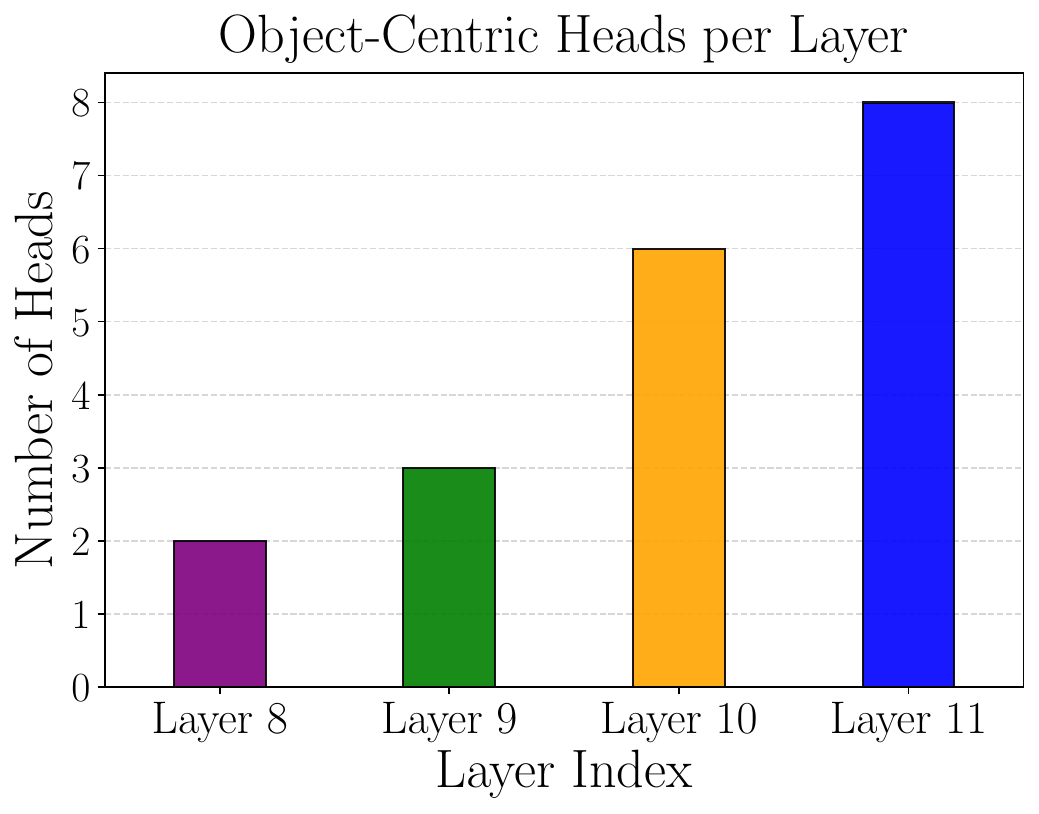}
                \caption*{(c) Highly active heads per layer.}
            \end{minipage}
        \end{tabular}
    }
    \vspace{-2pt}
    \caption{Analysis of the object-centric head distribution, computed over 4,000 images from the COCO dataset. (a) The heatmap shows the frequency of heads (across all layers) belonging to the object-centric cluster. The red boxes highlight that numerous heads in intermediate layers (e.g., Layers 8–10) are consistently selected, proving object-centric information is a distributed phenomenon and not confined to the final layer. (b) The plot details the final layer's contribution. It shows that several heads (circled in black) have a low frequency, confirming that some final-layer heads are "noisy" (non-object-centric). (c) The histogram shows the number of strongly active heads per layer.}
    \vspace{-5pt}
    \label{fig:analysis_figs}
\end{figure*}

To investigate this, we first propose a representation of object-centric behavior of the attention heads. We do this by combining the three matrices ($A_q$, $A_k$, $A_v$) into a unified object-centric ensemble matrix $A_{ens}$ (Eq.~\ref{eq:ensemble_sim}) for each head across the network. 
%
%
As stated before, we see (Fig.~\ref{fig:analysis_figs}) that the final layer indeed contains the highest concentration of object-centric heads. In addition, our analysis results in the following two key insights:
(1) some heads (4 out of 12, in Fig.~\ref{fig:analysis_figs} (b)) in the final layer are not object-centric and introduce noise that degrades localization quality when naively aggregated. (2) Numerous object-centric heads exist in intermediate layers (Layers 8-10 Fig.~\ref{fig:analysis_figs} (a)), containing valuable information, demonstrating that object discovery is not confined to the final layer, but distributed across the network.

Accordingly, we introduce Object-DINO, a training-free method that automatically discovers this distributed set of object-centric heads. The method operates by first characterizing every attention head across all layers using the ensemble similarity matrix $A_{ens}$ (Eq.~\ref{eq:ensemble_sim}), then clustering heads based on these ensembles. As discussed above, given the concentration of object-centric heads in the final layer, we select the cluster $c_{\text{obj}}$ with the highest proportion of final-layer heads. This yields a set $\mathcal{H}_{\text{obj}}$ of object-centric heads distributed across the network, while filtering out non-object-centric heads. Critically, our method requires no additional training, fine-tuning, or labeled data.

We validate Object-DINO's utility on two applications. First, integrating it into the TokenCut \cite{tokencut} framework yields significant performance gains for unsupervised object discovery, without any training. Second, we apply it to mitigate object hallucination in Multimodal Large Language Models (MLLMs) by providing visual evidence of object presence.

Our contributions are:

\begin{enumerate}
    \item We perform a systematic analysis of object-centric information in self-supervised ViTs, providing two key insights: (1) The information is encoded in the similarity matrices derived from all three attention components ($A_q, A_k, A_v$), not just key features or \texttt{[CLS]} attention used in the previous work, and (2) object-centric information is distributed across the network rather than confined to the final layer.
    \item Building on these findings, we introduce Object-DINO, a training-free method that automatically extracts this distributed object-centric information by clustering attention heads based on their ensemble matrices and identifying the cluster specializing in object-centric characteristics.
    \item We validate Object-DINO's effectiveness on two applications: (i) enhancing unsupervised object discovery with +3.6 to +12.4 CorLoc improvements across benchmarks, and (ii) mitigating object hallucination in MLLMs through visual grounding.
\end{enumerate}

\section{Related Works}
\label{sec:related_works}

\textbf{Emergent Localization in Self-Supervised ViTs.} Self-supervised learning (SSL) with Vision Transformers (ViTs) has demonstrated a remarkable ability to learn rich visual representations without human labels. Models like DINO \cite{dino} exhibit emergent properties, including semantic segmentation capabilities and object localization \cite{dino,amir2021deep}.  DINO's \texttt{[CLS]} token attention maps in the final layer often highlight salient objects, revealing an implicit understanding of scene structure. However, the \texttt{[CLS]} token is optimized for global image-level matching rather than localization, resulting in noisy maps that cannot be directly used for object discovery. This limitation motivates exploring alternatives to the \texttt{[CLS]} token for localization within the network.

\textbf{Attention Analysis and Head Selection in Vision Transformers.} Understanding the role of individual attention heads has been crucial for both interpreting and optimizing Vision Transformers. Early analysis in NLP revealed that different heads specialize in distinct linguistic functions \cite{voita2019analyzing,clark2019does}, inspiring similar investigations in vision. Recent vision works analyze head specialization on perception tasks \cite{dino,zhou2022understanding}, while others aggregate multi-head attention \cite{abnar2020quantifying}. However, these approaches typically analyze heads for computational efficiency \cite{michel2019sixteen,wang2019structured,liang2021pruning}, or apply post-hoc interpretability methods to task-specific, fine-tuned models \cite{abnar2020quantifying,chefer2021transformer}. While self-supervised methods like DINO \cite{dino} demonstrated that object-centric properties emerge during pre-training, this analysis was often qualitative and not the primary focus. In contrast, we perform a systematic, data-driven analysis of all heads in pre-trained self-supervised models. Our goal is to automatically identify and categorize the full spectrum of object-centric specializations, without relying on task-specific training or manual inspection.

\textbf{Unsupervised Object Discovery.}
Discovering objects without manual annotations has attracted significant attention with the emergence of self-supervised Vision Transformers. Early methods, such as DINO-seg \cite{dino}, applied simple thresholding to the \texttt{[CLS]} attention map. However, as discussed, the \texttt{[CLS]} token is noisy. More sophisticated approaches moved away from \texttt{[CLS]} and instead leveraged patch-level representations. They treat patches as nodes in a graph and apply spectral clustering or normalized min-cut \cite{tokencut,simeoni2021localizing}. TokenCut \cite{tokencut}, for instance, constructs its graph using only the key ($K$) features from the final layer. While this choice demonstrates improved performance over \texttt{[CLS]}-based methods, it raises two questions: (1) Why only keys ($K$)? Do queries($q$) and values($v$) contain complementary information? (2) Is the object-centric information confined to the final layer?
Our work systematically addresses these questions through a comprehensive analysis of all attention components across the network.

\begin{figure*}
    \centering
    \includegraphics[width=\linewidth]{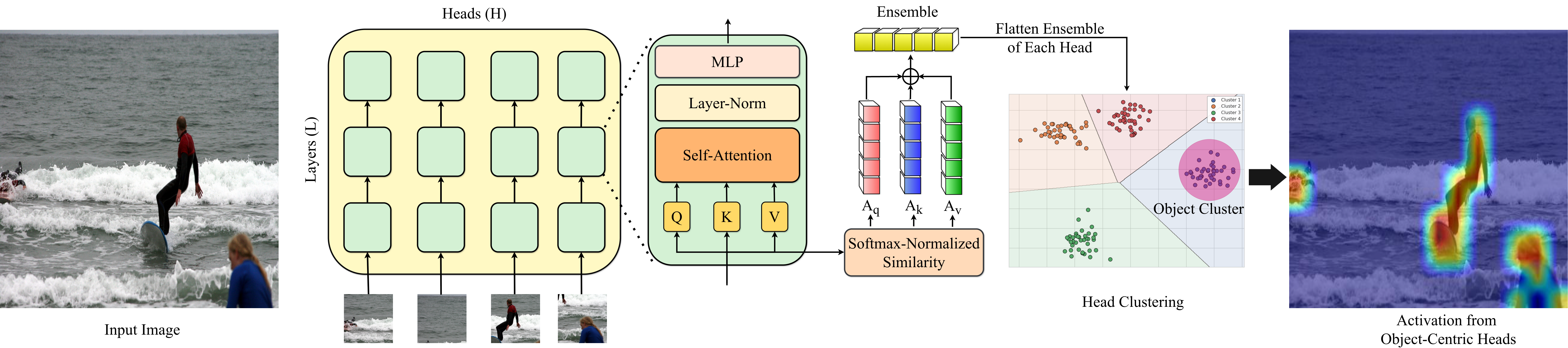}
    \caption{\textbf{Object-DINO Overview}. Our training-free algorithm identifies a distributed set of object-centric heads from a pre-trained models such as DINO. First, for every head across all layers, we compute the patch similarity maps from its query ($A_{q}$), key ($A_{k}$), and value ($A_{v}$) representations. Second, these three maps are ensembled and flattened to create a vector representing each head's localization pattern. Third, all heads are clustered based on these pattern. Fourth, guided by the established observation that object-centric information is most prevalent in the final layer \cite{dino,tokencut,simeoni2021localizing}, we automatically identify the object Cluster ($c_{\text{obj}}$) using the criterion that it contains the highest prevalence of final-layer heads. Finally, aggregating the similarity maps from only the object-centric heads ($\mathcal{H}_{\text{obj}}$) produces a high-fidelity object localization, effectively filtering noise from non-object-centric heads. For visualization, we invert the maps to show bright colors for objects.}
    \label{fig:main_figure}
\end{figure*}

\textbf{Object Hallucination in MLLMs.} A failure mode in MLLMs is object hallucination \cite{ji2023survey, liu2024survey}, where MLLMs generate incorrect information, often stemming from flawed visual grounding \cite{wang2023evaluation, shu2025large}. Mitigation strategies often intervene at inference time. One line of work aims to reduce over-reliance on the language prior through contrastive decoding: VCD \cite{vcd} contrasts logits from the original image against those from a noisy image to isolate the visual signal. M3ID \cite{m3id} amplifies the visual-conditioned signal over the unconditioned one, while other works use image augmentations \cite{ritual}. Another line of work leverages external models to provide explicit visual guidance. MARINE \cite{marine} uses supervised object detectors to generate a list of object candidates, which is then converted into text descriptions and used as a textual prompt to steer generation. Further approaches, like DeGF \cite{degf}, use a multi-pass feedback loop with a diffusion model. Our work also provides external guidance. Instead of relying on a multiple closed-set, supervised detector, we leverage the open-set, emergent objectness information from a single self-supervised model. This approach is not only more computationally efficient than running large detectors \cite{marine} or generative feedback loops \cite{degf}, but it also provides a more foundational grounding information. By extracting a direct visual-spatial map instead of a lossy text list, our method preserves spatial information.
\begin{algorithm}[t]
    \caption{Object-DINO: Head Clustering}
\label{alg:dino_spotlight_twophase}
\begin{algorithmic}[1]

\State \textbf{Input:} Pretrained ViT $\mathcal{M}$ (with $L$ layers, $H$ heads/layer), Dataset $\mathcal{D}$, Num clusters $K$
\State \textbf{Output:} Set of object-centric heads $\mathcal{H}_{\text{obj}}$
\Statex \vspace{0.1em}
\Statex \textbf{\textcolor{blue}{Phase 1: Head Feature Extraction}}
\State Images $\{x_i\}_{i=1}^N$ from $\mathcal{D}$

\For{each image $x_i$}
    \State Pass $x_i$ through $\mathcal{M}$ to get $q, k, v$ for all tokens
    \For{each layer $\ell \in [1, L]$ and head $h \in [1, H]$}
        \State Extract $r \in \{ q^{\ell,h}, k^{\ell,h}, v^{\ell,h} \}$ from $\mathcal{M}(x_i)$
        \State Normalize: $\tilde{r}^{\ell,h} = r^{\ell,h} / \|r^{\ell,h}\|$

        \State Compute self-similarity score
        \[
        A_r^{\ell,h} = \text{softmax}\!\left( \frac{ \tilde{r}^{\ell,h} \cdot (\tilde{r}^{\ell,h})^\top }{ \tau } \right)
        \]

        \State $A^{\ell,h}_{x_i} = \sum w_r A_r^{\ell,h},$ $ \forall \ r \in \{q,k,v\} $ \Comment{\textcolor{blue}{Ensemble}}
        
        \State $f^{\ell,h}_{x_i} = \text{Flatten}(A^{\ell,h}_{x_i})$ \Comment{\textcolor{blue}{Flatten map}}
    \EndFor
\Statex \vspace{0.1em}
    \Statex \textbf{\textcolor{blue}{Phase 2: Head Clustering and Selection}}
    \State Cluster $ f^{\ell,h}_{x_i}$ into K groups
    \State $C = \text{k-means}(f^{\ell,h}_{x_i}, K)$ \Comment{\textcolor{blue}{$C$ is a list of clusters}}
    
    \For{$k = 1$ to $K$}
        \State $n_k^{(L)} = |\{(\ell, h) \in \mathcal{C}_k \mid \ell = L\}|$ \Comment{\textcolor{blue}{Count heads from final layer in each cluster}}
    \EndFor
    \State $c_\text{obj} =\arg\max_{k} n_k^{(L)}$ \Comment{\textcolor{blue}{Cluster with most final-layer heads}}
    
    \State $\mathcal{H}_{\text{obj}} = \emptyset$
    \State $j = 0$ \Comment{\textcolor{blue}{Initialize flat index counter}}
    \For{each layer $\ell \in [1, L]$}
        \For{each head $h \in [1, H]$}
            \State $j = j + 1$ \Comment{\textcolor{blue}{Map $(\ell, h)$ to the list index $j$}}
            
            \If{\text{ClusterID}$[j] == c_{\text{obj}}$}
                \State $\mathcal{H}_{\text{obj}} = \mathcal{H}_{\text{obj}} \cup \{ (\ell, h) \}$
            \EndIf
        \EndFor
    \EndFor
    
    \State \Return $\mathcal{H}_{\text{obj}}$
\EndFor
\end{algorithmic}
\end{algorithm}

\section{Method}
\label{sec:method}

\subsection{Preliminaries}
\textbf{Vision Transformer.} A Vision Transformer (ViT) processes an image by first dividing it into a sequence of fixed-size patches. Each patch is linearly projected into an embedding vector. This sequence of patch embeddings is then fed through a series of Transformer encoder layers, wherein the multi-head self-attention (MHSA) mechanism allows every patch to dynamically exchange and aggregate information with every other patch in the sequence.

\noindent\textbf{Attention Components.} Within each self-attention head, the input embedding for every patch is transformed by three separate linear projections to create the \textbf{Query ($q$)}, \textbf{Key ($k$)}, and \textbf{Value ($v$)} vectors. These vectors are the fundamental components of the attention mechanism and serve distinct purposes:
\begin{itemize}
    \item \textbf{Query ($q$):} Represents what information a patch is seeking. It encodes the ``question'' a patch poses to identify relevant information from other patches.
    \item \textbf{Key ($k$):} Represents what information a patch offers. It acts as a descriptor of the patch's content that other patches' queries can match against.
    \item \textbf{Value ($v$):} Represents the actual content or features of the patch. This is the information that gets aggregated and propagated when the patch is attended to.
\end{itemize}

\noindent\textbf{Similarity Matrices of Attention Components.} For each attention component type $r \in \{q, k, v\}$ at layer $\ell$ and head $h$, we compute a normalized self-similarity matrix $A_r^{\ell,h}$ (Eq.~\ref{eq:self_sim}) in terms of dot products followed by softmax. These matrices provide complementary views of the image's internal structure:
\begin{itemize}
    \item \textbf{Query Similarity ($A_q$):} Identifies patches seeking similar information. High similarity between two patches indicates they perform similar information-seeking roles within that attention head.
    \item \textbf{Key Similarity ($A_k$):} Identifies patches offering similar contextual information. It groups patches that present themselves as having similar attributes or relevance.
    \item \textbf{Value Similarity ($A_v$):} Measures patches with similar content. It provides a direct comparison of the patches' feature representations.
\end{itemize}
\paragraph{}As stated in the introduction, our method is built upon two key insights:
(1) The patch-level similarity matrices derived from attention components ($A_q, A_k, A_v$) encode object-centric information. (2) The information is distributed across the network rather than confined to the final layer. They motivate Object-DINO, a training-free method that uses this distributed information to automatically identify and aggregate object-centric heads. We detail our findings and the algorithm in the following sections.

\subsection{Finding 1: Patch Similarity Matrices Encode Object Centric Information}
\label{sec:observation1}

We now describe the computation of $A_r^{\ell,h}$, which represents a head.
For any given head $(\ell, h)$ in the network and for each attention component  $r \in \{Q, K, V\}$ of the patches, we compute its patch self-similarity matrix $A_r^{\ell,h}$:
\begin{equation}
A_r^{\ell,h} = \text{softmax}\left( \frac{ \tilde{r}^{\ell,h} \cdot (\tilde{r}^{\ell,h})^\top }{ \tau } \right)
\label{eq:self_sim}
\end{equation}
where $\tilde{r}^{\ell,h} = r^{\ell,h} / \|r^{\ell,h}\|$ is the L2-norm of the representations, and $\tau$ is a temperature scaling factor. We use $\tau$ = 60 based on ablation studies (supplementary Sec. \ref{sec_supp:ablate_temp}).

While these individual maps ($A_q^{\ell,h}$, $A_k^{\ell,h}$, and $A_v^{\ell,h}$) all demonstrate localization properties, they capture different aspects of patch relationships and can be individually noisy. For instance, the $A_q$ and $A_v$ maps sometimes erroneously group background regions with the foreground, and $A_k$ misses out on a portion of Man (Fig.~\ref{fig:teaser}, top-row). To mitigate these artifacts and form a single map that leverages their complementary strengths, we ensemble them to create a unified object-centric map, $A_{ens}^{\ell,h}$:
\begin{equation}
\label{eq:ensemble_sim}
A^{\ell,h}_{ens} = w_q \cdot A_q^{\ell,h} + w_k \cdot A_k^{\ell,h} + w_v \cdot A_v^{\ell,h}
\end{equation}
Where, $w_q$, $w_k$, $w_v$ represents the scaling factors for each map. We use ($w_q = w_k = w_v$ = 0.33) in Eq. \ref{eq:ensemble_sim} (see supplementary Sec. \ref{supp_sec:weight_ablation} for weight ablation).
This ensemble map provides a low-noise representation of object saliency. We empirically justify the robustness of this ensemble over the individual $A_q, A_k, \text{and } A_v$ maps in Fig. \ref{fig:qkv_ensemeble_head_selection}. We use $A_{ens}^{\ell,h}$ as the unified object-centric map to characterize and analyze every head in the network.

\subsection{Finding 2: Object-Centricity is a Distributed Property}
\label{sec:observation2}    
We now investigate how $A_r^{\ell,h}$ is distributed across the network. Prior work \cite{dino,tokencut} has primarily focused on the final layer, but whether object-centric heads are confined there or distributed throughout the network has not been considered. Similarly, it is unclear whether all final-layer heads contribute equally to object discovery.

To answer this, we perform analysis on 4,000 random images from the COCO dataset \cite{coco}. We cluster the ensemble maps $A_{ens}^{\ell,h}$, using the clustering method (Sec.~\ref{sec:dino_spotlight}), which allows us to group heads based on their function. Our findings, summarized in Fig.~\ref{fig:analysis_figs}, are:
\begin{enumerate}
    \item \textbf{Object-centric heads are not exclusive to the final layer.} We find that the set of object-centric heads ($\mathcal{H}_{\text{obj}}$ ) consistently contains numerous heads from intermediate and even early layers.
    \item \textbf{Not all final-layer heads are object-centric.} Conversely, a few heads from the final layer ($L$) are consistently assigned to non-object clusters, implying that they contribute noise when used for object discovery.  Supplementary Sec.\ref{supp_sec:head_viz} visualizes the ensemble map for heads in final layer.
\end{enumerate}

\noindent These results demonstrate that simply selecting all heads as object-centric from the final layer is a suboptimal strategy. 

\begin{figure}
    \centering
    \includegraphics[width=\linewidth]{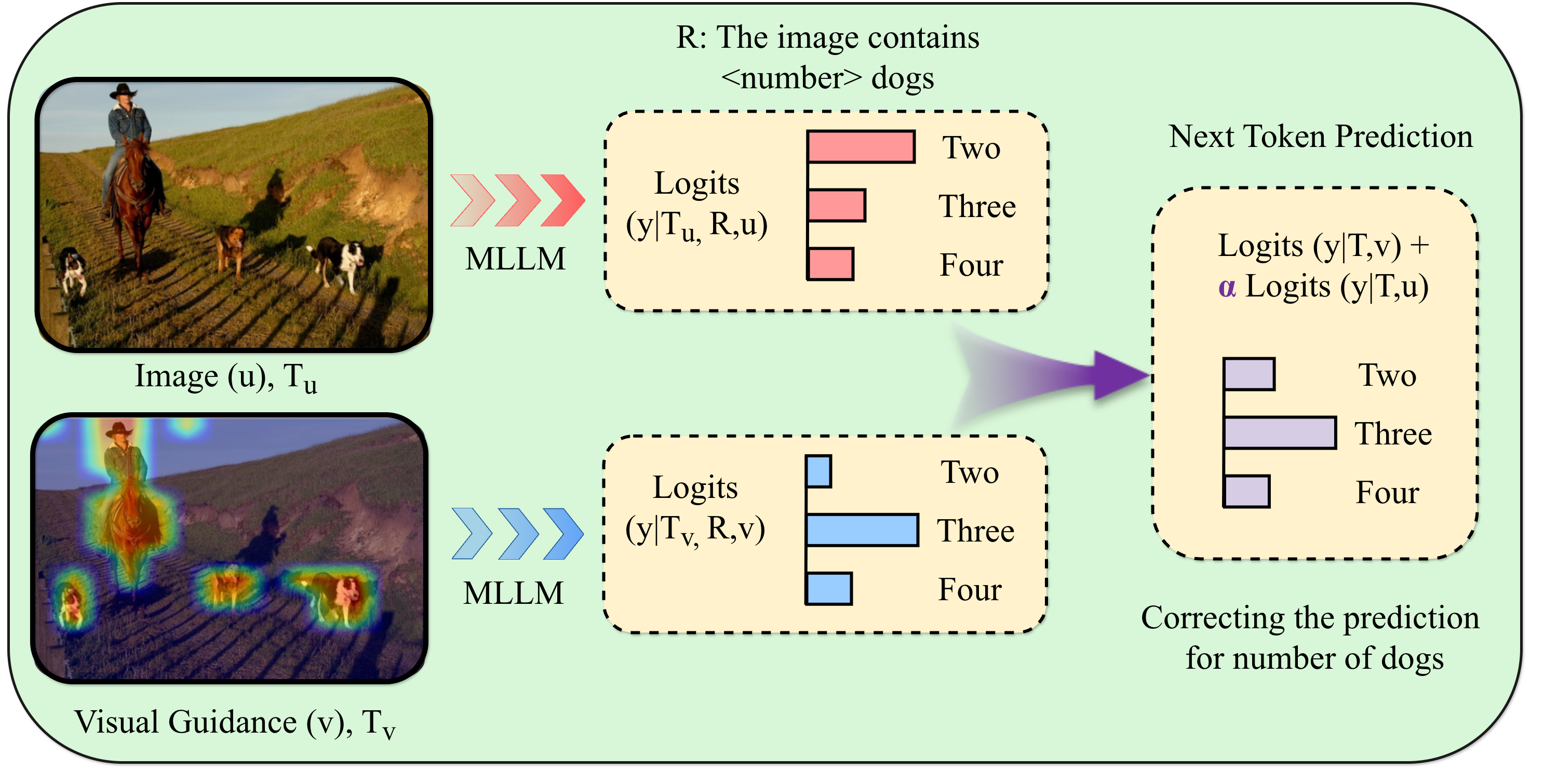}
    \caption{\textbf{Mitigating MLLM Hallucination with Object-DINO Visual Guidance.} Our training-free decoding strategy computes two separate logit distributions. The standard branch uses the original image $u$ and a general prompt $T_u$ (e.g., "describe this image"). The guidance branch uses the Object-DINO's object map $v$ and a prompt $T_v$ (e.g., "describe the highlighted regions"). Here, R is the text prompt generated so far. We then add these logits, $Logits(y|T_u, R, u) + \alpha \cdot Logits(y|T_v, R, v)$, to amplify tokens consistent with the visual evidence, thus correcting the hallucinations (e.g., "Two" $\rightarrow$ "Three" dogs)}
    \label{fig:MLLM_fig}
\end{figure}

\subsection{Object-DINO: Algorithm for Head Selection} \label{sec:dino_spotlight}

We now present our algorithm, Object-DINO that automatically selects the set of object-centric heads ($\mathcal{H}_{\text{obj}}$), without any training.
The algorithm operates in two phases.

\noindent{\textbf{Phase 1: Head Feature Extraction.}}
To identify the heads responsible for localization, we must first characterize the behavior of every head $(\ell, h)$. For each image $x_i$ in a dataset $\mathcal{D}$, we compute the ensemble map $A^{\ell,h}_{x_i}$ from Eq.~\ref{eq:ensemble_sim}. As specified in the algorithm, this map is then flattened to create a feature vector $f^{\ell,h}_{x_i}$ that describes the head's behavior on that specific image. Since heads that specialize in similar functions (e.g., finding objects) produce correlated patterns \cite{voita2019analyzing}, their feature vectors will naturally cluster together. This property is what allows us to isolate the object-centric group via unsupervised clustering.
\\

\noindent{\textbf{Phase 2: Head Clustering and Selection.}} Following this logic, this phase (Algorithm~\ref{alg:dino_spotlight_twophase}) analyzes the flattened features ($f^{\ell,h}_{x_i}$) to find the network-wide set of object-centric heads. For each image, we have $L \times H$  flattened features ($f_{x_i}$) describing the heads of each layer. Next, we apply k-means clustering to partition all $L \times H$ heads into $K$ distinct groups based on their ($f^{\ell,h}_{x_i}$)
\begin{equation}
\begin{aligned}
     C &= \text{k-means}(f^{\ell,h}_{x_i}, K), \\
    \text{where} \quad C &= \{C_1, \dots, C_K\} \text{ are the clusters.}
\end{aligned}
\end{equation}
Once the clusters are formed with $K$=5 clusters (see supplementary Sec. \ref{supp_sec:optimal_k}), we must automatically identify which cluster contains object-centric heads. Given that object-centric information is most concentrated in the final layer (as shown in our analysis and prior work~\cite{dino,tokencut}), we select the cluster $c_{\text{obj}}$ with the highest number of final-layer heads as our object-centric cluster. Our analysis in Fig.~\ref{fig:analysis_figs}(c) validates this strategy:  while the final layer (Layer 11) indeed shows the highest concentration of object-centric heads, numerous heads from intermediate layers (8-10) are also selected, confirming that object-centric information is distributed across the network rather than confined to a single layer.
\begin{equation}
\begin{aligned}
    c_\text{obj} &= \arg\max_{k} n_k^{(L)}, \\
    \text{where} \quad n_k^{(L)} &= |\{(\ell, h) \in \mathcal{C}_k \mid \ell = L\}|
\end{aligned}
\end{equation}
The object-centric heads ($\mathcal{H}_{\text{obj}}$ ) are those that are assigned to this single cluster, $\mathcal{H}_{\text{obj}} = \{(\ell, h) \mid (\ell,h) \in c_{\text{obj}}\}$.
%
%
%
%
%
%
%
%
%
%
%
\begin{table}[tp]
\captionof{table}{\textbf{Unsupervised Object Discovery.} We evaluate Object-DINO by integrating it into the TokenCut \cite{tokencut} framework. We replace TokenCut's baseline strategy (using all final-layer heads) with our automatically discovered set of object-centric heads. This change yields substantial performance gains (in \textcolor{blue!90!black}{blue}) for both DINO-V2 and DINO-V3 models across all datasets.}
\label{tab:tokencut}
\resizebox{\columnwidth}{!}{
\begin{tabular}{l l ccc}
\toprule
\textbf{Model} & \textbf{Method} & \textbf{VOC07} & \textbf{VOC12} & \textbf{COCO20K} \\
\midrule
\multirow{2}{*}{Dino-V3 \cite{dinov3}} 
& \cellcolor{lightgray!45}TokenCut \cite{tokencut} & \cellcolor{lightgray!45}26 & \cellcolor{lightgray!45}30.3 & \cellcolor{lightgray!45}19.8 \\
& + Ours & 30.8 {\scriptsize (\textcolor{blue!90!black}{\textbf{$\uparrow$4.8}})} 
             & 36.0 {\scriptsize (\textcolor{blue!90!black}{\textbf{$\uparrow$5.7}})} 
             & 23.4 {\scriptsize (\textcolor{blue!90!black}{\textbf{$\uparrow$3.6}})} \\
\midrule
\multirow{2}{*}{Dino-V2 \cite{dinov2}} 
& \cellcolor{lightgray!45} TokenCut \cite{tokencut} & \cellcolor{lightgray!45} 16.2 & \cellcolor{lightgray!45} 18.3 &  \cellcolor{lightgray!45} 11.9 \\
& + Ours & 25.7 {\scriptsize (\textcolor{blue!90!black}{\textbf{$\uparrow$9.5}})} 
             & 30.7 {\scriptsize (\textcolor{blue!90!black}{\textbf{$\uparrow$12.4})} }
             & 19.7 {\scriptsize (\textcolor{blue!90!black}{\textbf{$\uparrow$7.8}})} \\
\bottomrule
\end{tabular}

}
\end{table}

\section{Applications}
We now present the use of Object-DINO for two tasks: (1) Unsupervised object discovery and (2) Mitigate object hallucination in MLLMs.
\subsection{Unsupervised Object Discovery}
We demonstrate the effectiveness of Object-DINO on the task of unsupervised object discovery, which aims to localize objects without manual annotations such as segmentation masks or bounding boxes.

\textbf{Methodology.} We adopt the non-parametric, graph-based TokenCut algorithm \cite{tokencut} as our base framework. The standard TokenCut implementation aggregates the key ($k$) features from all attention heads in the final layer of a pre-trained DINO model \cite{dinov2,dinov3}, which ignores both valuable object-centric heads in intermediate layers and includes noisy, non-object-centric heads from the final layer (Sec. \ref{sec:observation2}). We validate our method by replacing this naive aggregation. We use Object-DINO (Sec.~\ref{sec:dino_spotlight}) to build a Token-cut like patch-affinity graph using ensemble-similarity of object-centric heads $\mathcal{H}_{\text{obj}}$ as affinity.

\textbf{Evaluation.} We conduct experiments on three standard benchmarks: PASCAL VOC 2007 \cite{pascal-voc}, VOC 2012 \cite{pascal-voc}, and COCO 20k \cite{coco20k}. Following the established protocol \cite{simeoni2021localizing,tokencut}, we report the CorLoc metric, which measures the percentage of images where the predicted bounding box has an IoU $> 0.5$ with a ground-truth box.

\textbf{Results.} As presented in Tab.~\ref{tab:tokencut}, our Object-DINO selection provides significant and consistent performance gains. When applied to DINO-V3 \cite{dinov3}, our method improves CorLoc by +4.8, +5.7, and +3.6 points on VOC 2007, VOC 2012, and COCO 20k, respectively. For DINO-V2 \cite{dinov2}, we obtain gains of +9.5, +12.4, and +7.8 points over the TokenCut baseline. These results demonstrate that our head selection method is effective and improves performance across different model versions.

\begin{figure}
    \centering
    \includegraphics[width=\linewidth]{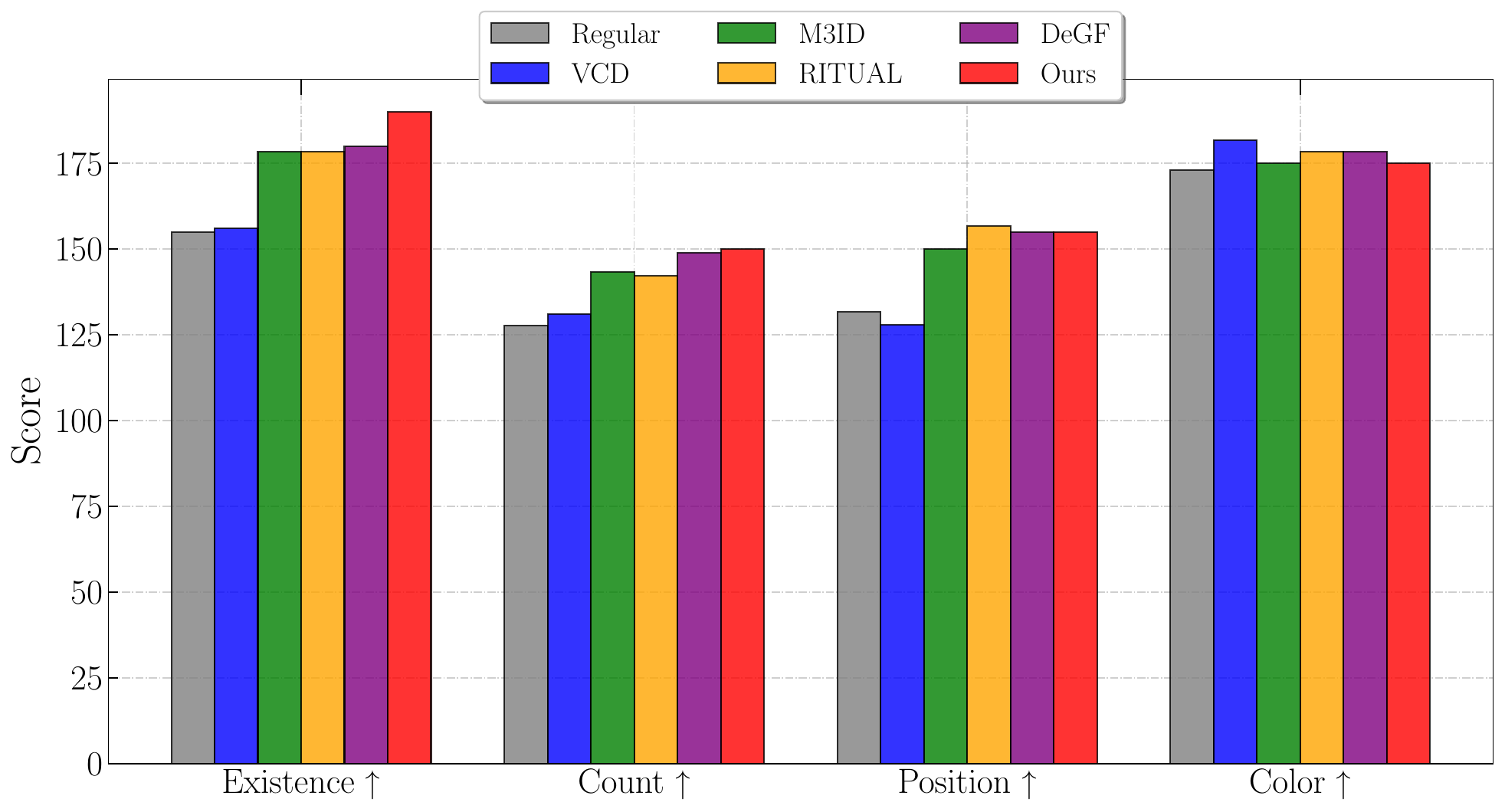}
    \caption{Experimental results of MME on the hallucination subset}
    \label{fig:MME_Qwen}
\end{figure}

\subsection{Mitigating MLLM Hallucination}
\label{sec:mllm_hall}
Our second application targets object hallucination in Multimodal Large Language Models (MLLMs), where models describe content not present in the image. Using Object-DINO, we provide explicit visual grounding to reduce such hallucinations.

\textbf{Methodology.} We propose a training-free decoding strategy that amplifies visual evidence during generation. As shown in Fig. \ref{fig:MLLM_fig}, we use two branches to compute the logits for the next token. In the standard branch, the MLLM takes the original image ($u$) and a prompt ($T_u$, e.g., "Describe this image."). At each decoding step, given the running text $R$ (the text generated so far), it produces logits $\text{Logits}(y|T_u, R, u)$ for the next token. The guidance branch uses Object-DINO to generate an object-centric map ($v$) highlighting visually grounded regions. This map ($v$) is then fed into the MLLM, along with a specific prompt ($T_v$, e.g., "Describe the highlighted regions."). The model is then prompted with $T_v$ (e.g., "Describe the highlighted regions."), producing guidance logits $\text{Logits}(y|T_v, R, v)$ that emphasize grounded evidence.

As in Eq. \ref{eq:mllm_combine}, we compute the logit distribution for the next token by linearly combining the guidance logits and the standard logits using the hyperparameter $\alpha$. We set $\alpha$ = 0.4 based on empirical evaluation (supplementary Sec. \ref{supp_sec:alpha_ablation}). This strategy effectively focuses on tokens that are consistent with the visual evidence, correcting potential hallucinations and enhancing the MLLM's faithfulness to the visual input.
\begin{equation} 
\label{eq:mllm_combine} L =\alpha \ \text{Logits}(y|T_u, R, u) + (1-\alpha)\text{Logits}(y|T_v, R, v) 
\end{equation}
\textbf{Evaluation.} Following prior works, we evaluate our method on three standard hallucination benchmarks: CHAIR \cite{chair}, POPE \cite{POPE}, and MME \cite{mme_data}. We apply our decoding strategy to several pre-trained MLLMs, including LLaVA-1.5 \cite{llava_1.5}, Instruct-BLIP \cite{instruct_blip}, and Qwen-VL \cite{qwen}.

\textbf{Results.} On the POPE benchmark (Tab.~\ref{tab:POPE}), our method achieves the highest Precision and F1-score across all three MLLMs: LLaVA-1.5, Instruct-BLIP, and Qwen-VL. While it attains the second-best Accuracy on LLaVA-1.5, it consistently delivers the strongest Precision and F1, providing the most reliable hallucination mitigation overall. For the CHAIR benchmark (Tab.~\ref{tab:CHAIR}), our method substantially reduces hallucination errors. It obtains the lowest Cs and Ci scores on LLaVA-1.5 and the lowest Cs score on Instruct-BLIP, while among the lowest Ci result among prior methods.
Finally, our method achieves the best performance on the MME hallucination subset across majority of the categories(Fig.~\ref{fig:MME_Qwen}). Additionally, our method remains highly efficient. As shown in the supplementary (Sec.~\ref{Supp_sec:MLLM_efficiency}) and Tab.~\ref{tab:latency}, it introduces only a modest increase in latency and memory compared to the regular baseline and VCD decoding, while achieving substantially lower hallucination rates. By contrast, advanced decoding methods with multi-stage feedback loops incur significantly higher computational overhead, making our approach both faster and more memory-efficient. These results show the applicability of using Object-DINO as explicit visual evidence to guide MLLM generation.
\begin{table}[tp]
\captionof{table}{\textbf{POPE Results.} Results on POPE benchmark. Higher accuracy, precision, and F1 indicate better performance. The best results in bold, and the second-best are underlined.}
\label{tab:POPE}
\resizebox{\columnwidth}{!}{

\begin{tabular}{lccccccccc}
\toprule
\textbf{Method} &
\multicolumn{3}{c}{\textbf{LLaVA-1.5 \cite{llava_1.5}}} &
\multicolumn{3}{c}{\textbf{Instruct-BLIP \cite{instruct_blip}}} &
\multicolumn{3}{c}{\textbf{Qwen-VL \cite{qwen}}} \\
\cmidrule(lr){2-4} \cmidrule(lr){5-7} \cmidrule(lr){8-10}
 & \textbf{Acc} & \textbf{P} & \textbf{F1} & \textbf{Acc} & \textbf{P} & \textbf{F1} & \textbf{Acc} & \textbf{P} & \textbf{F1} \\
\midrule
\rowcolor{lightgray!45} Regular & 77.4 & 73.3 & 79.2 & 74.6 & 71.2 & 76.4 & 79.8 & 80.1 & 79.7 \\
VCD  & 77.1 & 72.1 & 79.4 & 77.2 & 74.2 & 78.4 & 81.3 & 80.6 & 81.5 \\
\rowcolor{lightgray!45} M3ID & 78.2 & 73.5 & 80.2 & 77.4 & 73.6 & 79.1 & 82.0 & 81.4 & 82.1 \\
RITUAL & 78.8 & 74.4 & 80.5 & 78.7 & 74.5 & \underline{80.3} & 82.8 & 83.1 & 82.7 \\
\rowcolor{lightgray!45} MARINE & \textbf{83.9} & - & 82.7 & 79.13 & - & 77.1 & - & - & - \\
 DeGF & 81.6 & \underline{80.5} & \underline{81.9} & 
\underline{80.3} & \underline{80.9} & 80.1 & \underline{83.4} & \underline{84.4} & \underline{82.9} \\

\midrule
\rowcolor{orange!35} \textbf{Ours} & \underline{83.6} & \textbf{87.4} & \textbf{82.7} & \textbf{82.7} & \textbf{87.7} & \textbf{81.6} & \textbf{86.6} & \textbf{89.2} & \textbf{86.1} \\
\bottomrule
\end{tabular}

}
\end{table}
\begin{table}
\small
\captionof{table}{\textbf{CHAIR Results.} Results on CHAIR benchmark with the maximum number of new tokens set to 64. Lower CHAIRs and CHAIRi (Ci) indicate less hallucination. Best results are in bold, second-best are underlined.}
\label{tab:CHAIR}
\centering
\resizebox{0.81\columnwidth}{!}{
\begin{tabular}{lcccc}
\toprule
\textbf{Method} & \multicolumn{2}{c}{\textbf{LLaVA-1.5 \cite{llava_1.5}}} & \multicolumn{2}{c}{\textbf{Instruct-BLIP \cite{instruct_blip}}} \\
\cmidrule(lr){2-3} \cmidrule(lr){4-5}
 & \textbf{Cs} & \textbf{Ci} & \textbf{Cs} & \textbf{Ci} \\
\midrule
\rowcolor{lightgray!45} Regular         & 26.2 & 9.4 & 31.2 & 11.1 \\
VCD \cite{vcd}             & 24.4 & 7.9 & 30.0 & 10.1 \\
\rowcolor{lightgray!45} M3ID \cite{m3id}           & 21.4 & 6.3 & 30.8 & 10.4 \\
RITUAL \cite{ritual}           & 22.4 & 6.9 & 26.6 & 8.9  \\
\rowcolor{lightgray!45} Woodpecker \cite{woodpecker}      & 24.9 & 7.5 & 31.2 & 10.8 \\
HALC \cite{halc}             & 21.7 & 7.1 & 24.5 & \underline{8.0}  \\
\rowcolor{lightgray!45}MARINE \cite{marine}           & 20.8 & \underline{6.0}& 27.5 & 8.8 \\
 DeGF \cite{degf}           & \textbf{18.4} & 6.1 & \underline{24.0} & \textbf{7.7}  \\
\midrule
\rowcolor{orange!35} \textbf{Ours} & \textbf{18.4} & \textbf{5.9} & \textbf{21.4} & 8.8  \\
\bottomrule
\end{tabular}

}
\end{table}
\begin{table}[tp]
\captionof{table}{\textbf{Layer Head Ablation.} This table validates our two hypotheses. (1) \textbf{Row-2}: Selecting only object-centric heads from the final layer improves over the TokenCut baseline (+1.5 CorLoc on VOC07), confirming that some final-layer heads contribute noise. (2) \textbf{Row-3}: Selecting object-centric heads from all layers leads to further gains (+3.3 additional CorLoc on VOC07), proving that object-centric information is distributed across intermediate layers and accounts for the majority of the performance improvement.}
\label{tab:layer_head_ablation}
\resizebox{\columnwidth}{!}{

\begin{tabular}{lccc}
\toprule
\textbf{Method} & \textbf{VOC07} & \textbf{VOC12} & \textbf{COCO20K} \\
\midrule
 TokenCut \cite{tokencut} & 26.0 & 30.3 & 19.8 \\
\rowcolor{lightgray!45}\textit{+ Our Head (final layer)} & 27.5 {\scriptsize (\textcolor{blue!90!black}{\textbf{$\uparrow$1.5}})} & 31.4 {\scriptsize (\textcolor{blue!90!black}{\textbf{$\uparrow$1.1}})} & 20.5 {\scriptsize (\textcolor{blue!90!black}{\textbf{$\uparrow$0.7}})} \\
 \textit{+ Our Head (all layers)} & 30.8 {\scriptsize (\textcolor{blue!90!black}{\textbf{$\uparrow$4.8}})} & 36.0 {\scriptsize (\textcolor{blue!90!black}{\textbf{$\uparrow$5.7}})} & 23.4 {\scriptsize (\textcolor{blue!90!black}{\textbf{$\uparrow$3.6}})} \\
\bottomrule
\end{tabular}

}
\end{table}
\section{Analysis}
 \textbf{Impact of Object-DINO Head Selection.} As shown in Tab.~\ref{tab:layer_head_ablation}, our method's advantages are twofold, directly validating our hypotheses. First, simply replacing TokenCut's features with features from our selected final-layer heads yields a consistent improvement (+1.5, +1.1, and +0.7 CorLoc on VOC07, VOC12, and COCO20k respectively), confirming that some final-layer heads are noisy. Second, using our selected heads from \emph{all} layers provides substantially larger gains, achieving gains of +4.8, +5.7, and +3.6 CorLoc on VOC07, VOC12, and COCO20k. Notably, intermediate 
layer heads contribute +3.3, +4.6, and +2.9 additional CorLoc beyond 
final-layer-only selection. This result confirms our findings: object-centric information is a distributed property, and crucial heads from earlier layers are ignored by final-layer-only approaches.
\\

\begin{figure}
    \centering
    \includegraphics[width=0.99\linewidth]{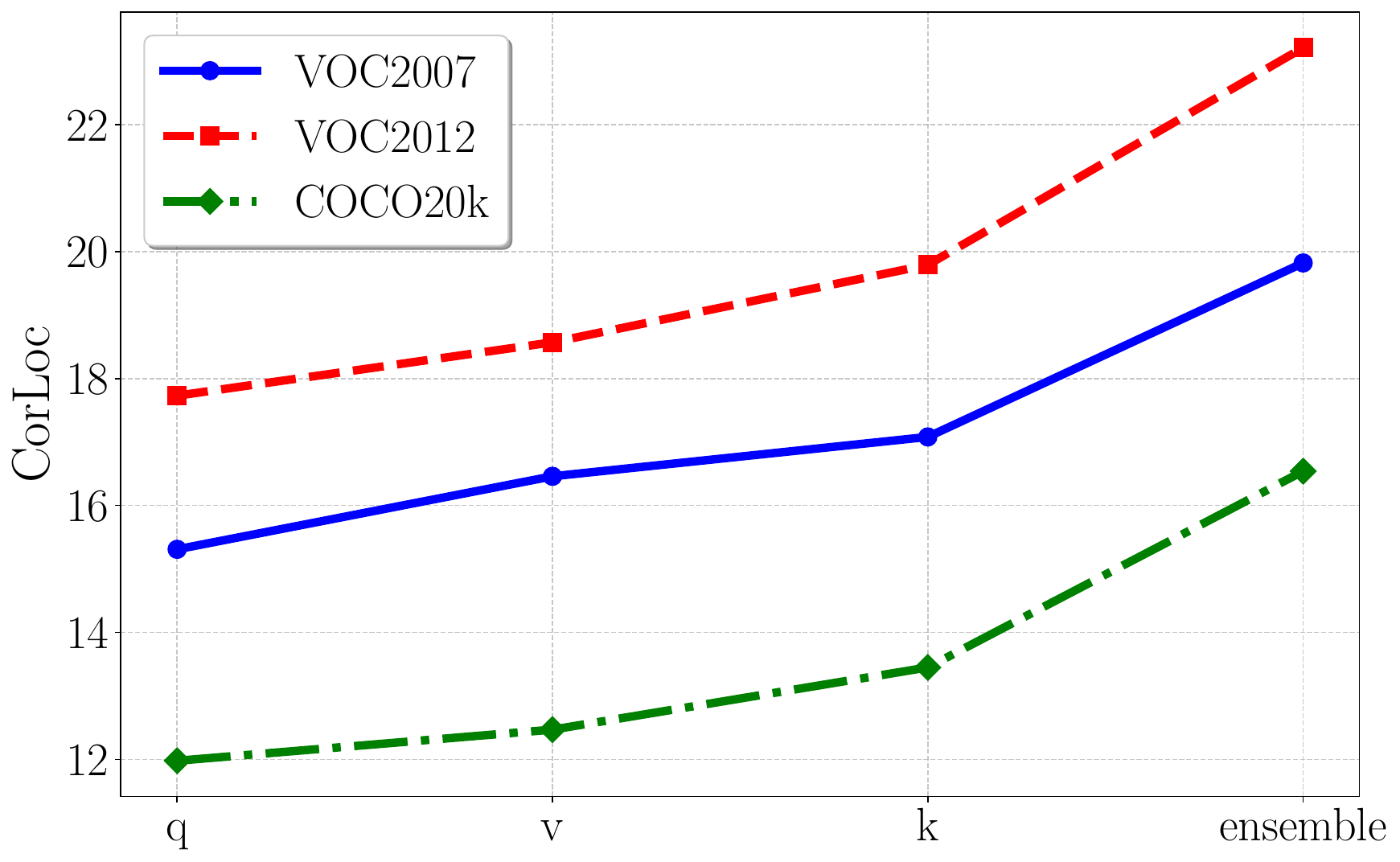}
    \caption{\textbf{Impact of Attention Components on Head Selection}. CorLoc performance across three datasets (VOC2007, VOC2012, COCO20k) when using different attention components ($Q, K, V$) versus their ensemble to identify object-centric heads via Object-DINO. The performance ordering is $Q < V < K < \text{Ensemble}$ across all datasets. This demonstrates that $\text{Ensemble}$ leads to identification of robust object-centric heads.}
    \label{fig:qkv_ensemeble_head_selection}
\end{figure}

\noindent\textbf{Attention Component for Head Selection.} We conduct an ablation to justify our use of the ensemble ($A_{ens}$) for head selection. We compare the downstream performance of Object-DINO when using the ensemble versus using each of the individual components ($A_q$, $A_k$, or $A_v$) as the head clustering feature. The evaluation is performed using the TokenCut framework on the VOC 2007, VOC 2012, and COCO 20k benchmarks. As shown in Fig.~\ref{fig:qkv_ensemeble_head_selection}, the results show a consistent performance ranking: $A_q < A_v < A_k < A_{ens}$ across all datasets. The ensemble consistently outperforms all the individual components, thus we use the ensemble similarity ($A_{ens}$) for head characterization.
\\

\noindent\textbf{Model Architecture and Objective.} In this section, we extend our analysis to: (1) larger DINO models (ViT-L), and (2) alternative self-supervised methods (MAE \cite{mae}). Applying our method to DINO ViT-L/14 on 4,000 COCO images confirms the distributed pattern persists across model scales (Supplementary Fig. \ref{supp_fig:analysis_figs}). For a MAE-ViT-B model to test the reconstruction-based pre-training objective. This analysis (Supplementary, Fig.\ref{fig:analysis_figs}) confirms that MAE also distributes object-centric information. Further, we observe that MAE's signals are significantly noisier (see Fig.\ref{fig:mae_vis}) than DINO's. This observation justifies the literature's (and our) focus on DINO for high-fidelity localization tasks.
\begin{figure}
    \centering
    \includegraphics[width=\linewidth]{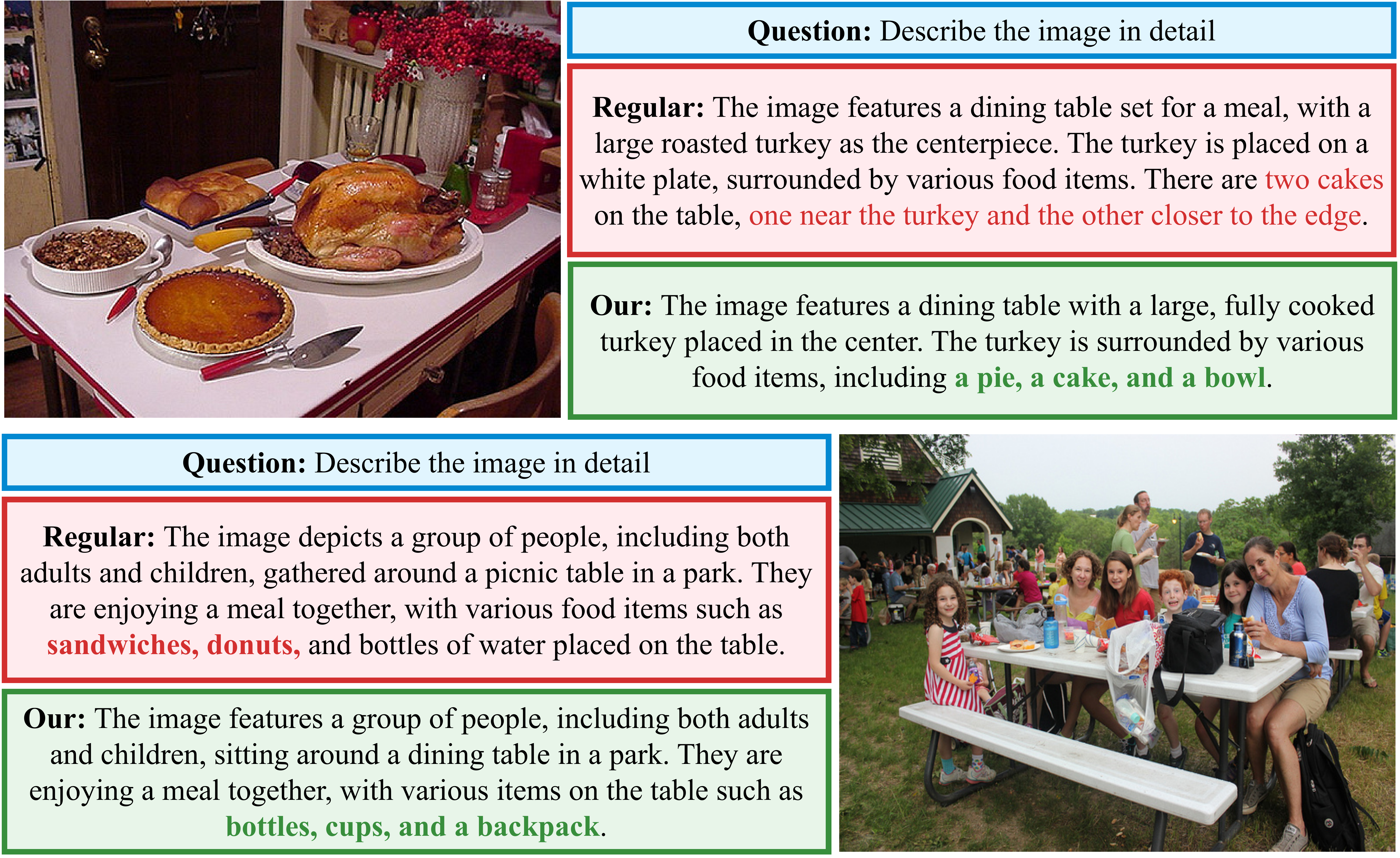}
    \caption{Qualitative comparison of captions generated by LLava using regular decoding (red box) vs our method (green box). }
    \label{fig:Hallucination_example}
\end{figure}

\section{Discussion}
We analyzed object-centric information in self-supervised Vision Transformers, revealing that this information is encoded across all attention components ($Q, K, V$), and distributed throughout the network rather than confined to the final layer. Our findings enabled us to develop Object-DINO, a training-free method that automatically discovers object-centric attention heads and significantly improves unsupervised object discovery and mitigates hallucinations in MLLMs.
%
%

\textbf{Limitations and Future Works.} While our approach effectively identifies the distributed objectness information, it is currently limited to producing a foreground-background separation. It does not perform accurate instance-level segmentation, which would require not only detecting object-centric regions but also decomposing them into an unknown number of distinct instances. Addressing this challenge is a natural extension of our work. Further, analyzing patch-level attention components in vision-language models (CLIP) can reveal whether distributed object-centric information emerges across different modalities.
%

\section{Acknowledgement}
The support of the Office of Naval Research under grant N00014-24-1-2169, Amazon–Illinois Center on AI for Interactive Conversational Experiences, USDA National Institute of Food and Agriculture under grant AFRI 2020-67021-32799/1024178, computational resources from UIUC NCSA-Delta under NSF award OAC-2005572, U.S. National Science Foundation (NSF) Grant CRII 2451683, an NVIDIA Academic Grants Program, a U.S. Bank Academic Research Award, the University of California, Merced, a UC Merced Faculty Research Award, Institute of Information \& Communications Technology Planning \& Evaluation (IITP) grant funded by the Korean Government (MSIT) (No. RS-2024-00457882, National AI Research Lab Project) are gratefully acknowledged. The views and conclusions are those of the authors and do not necessarily reflect the official policy or position of the U.S. Government.

{
    \small
    \bibliographystyle{ieeenat_fullname}
    \bibliography{main}

\begin{thebibliography}{35}
\providecommand{\natexlab}[1]{#1}
\providecommand{\url}[1]{\texttt{#1}}
\expandafter\ifx\csname urlstyle\endcsname\relax
  \providecommand{\doi}[1]{doi: #1}\else
  \providecommand{\doi}{doi: \begingroup \urlstyle{rm}\Url}\fi

\bibitem[Abnar and Zuidema(2020)]{abnar2020quantifying}
Samira Abnar and Willem Zuidema.
\newblock Quantifying attention flow in transformers.
\newblock \emph{arXiv preprint arXiv:2005.00928}, 2020.

\bibitem[Amir et~al.(2021)Amir, Gandelsman, Bagon, and Dekel]{amir2021deep}
Shir Amir, Yossi Gandelsman, Shai Bagon, and Tali Dekel.
\newblock Deep vit features as dense visual descriptors.
\newblock \emph{arXiv preprint arXiv:2112.05814}, 2\penalty0 (3):\penalty0 4, 2021.

\bibitem[Bai et~al.(2023)Bai, Bai, Chu, Cui, Dang, Deng, Fan, Ge, Han, Huang, et~al.]{qwen}
Jinze Bai, Shuai Bai, Yunfei Chu, Zeyu Cui, Kai Dang, Xiaodong Deng, Yang Fan, Wenbin Ge, Yu Han, Fei Huang, et~al.
\newblock Qwen technical report.
\newblock \emph{arXiv preprint arXiv:2309.16609}, 2023.

\bibitem[Caron et~al.(2021)Caron, Touvron, Misra, J{\'e}gou, Mairal, Bojanowski, and Joulin]{dino}
Mathilde Caron, Hugo Touvron, Ishan Misra, Herv{\'e} J{\'e}gou, Julien Mairal, Piotr Bojanowski, and Armand Joulin.
\newblock Emerging properties in self-supervised vision transformers.
\newblock In \emph{CVPR}, pages 9650--9660, 2021.

\bibitem[Chefer et~al.(2021)Chefer, Gur, and Wolf]{chefer2021transformer}
Hila Chefer, Shir Gur, and Lior Wolf.
\newblock Transformer interpretability beyond attention visualization.
\newblock In \emph{CVPR}, pages 782--791, 2021.

\bibitem[Chen et~al.(2024)Chen, Zhao, Luo, Yao, Li, and Zhou]{halc}
Zhaorun Chen, Zhuokai Zhao, Hongyin Luo, Huaxiu Yao, Bo Li, and Jiawei Zhou.
\newblock Halc: Object hallucination reduction via adaptive focal-contrast decoding.
\newblock \emph{arXiv preprint arXiv:2403.00425}, 2024.

\bibitem[Clark et~al.(2019)Clark, Khandelwal, Levy, and Manning]{clark2019does}
Kevin Clark, Urvashi Khandelwal, Omer Levy, and Christopher~D Manning.
\newblock What does bert look at? an analysis of bert's attention.
\newblock \emph{arXiv preprint arXiv:1906.04341}, 2019.

\bibitem[Dai et~al.(2023)Dai, Li, LI, Tiong, Zhao, Wang, Li, Fung, and Hoi]{instruct_blip}
Wenliang Dai, Junnan Li, DONGXU LI, Anthony Tiong, Junqi Zhao, Weisheng Wang, Boyang Li, Pascale~N Fung, and Steven Hoi.
\newblock Instructblip: Towards general-purpose vision-language models with instruction tuning.
\newblock In \emph{NeurIPS}, pages 49250--49267. Curran Associates, Inc., 2023.

\bibitem[Everingham et~al.(2010)Everingham, Van~Gool, Williams, Winn, and Zisserman]{pascal-voc}
Mark Everingham, Luc Van~Gool, Christopher~KI Williams, John Winn, and Andrew Zisserman.
\newblock The pascal visual object classes (voc) challenge.
\newblock \emph{IJCV}, 88:\penalty0 303--338, 2010.

\bibitem[Favero et~al.(2024)Favero, Zancato, Trager, Choudhary, Perera, Achille, Swaminathan, and Soatto]{m3id}
Alessandro Favero, Luca Zancato, Matthew Trager, Siddharth Choudhary, Pramuditha Perera, Alessandro Achille, Ashwin Swaminathan, and Stefano Soatto.
\newblock Multi-modal hallucination control by visual information grounding.
\newblock In \emph{CVPR}, pages 14303--14312, 2024.

\bibitem[He et~al.(2022)He, Chen, Xie, Li, Doll{\'a}r, and Girshick]{mae}
Kaiming He, Xinlei Chen, Saining Xie, Yanghao Li, Piotr Doll{\'a}r, and Ross Girshick.
\newblock Masked autoencoders are scalable vision learners.
\newblock In \emph{CVPR}, pages 16000--16009, 2022.

\bibitem[Ji et~al.(2023)Ji, Lee, Frieske, Yu, Su, Xu, Ishii, Bang, Madotto, and Fung]{ji2023survey}
Ziwei Ji, Nayeon Lee, Rita Frieske, Tiezheng Yu, Dan Su, Yan Xu, Etsuko Ishii, Ye~Jin Bang, Andrea Madotto, and Pascale Fung.
\newblock Survey of hallucination in natural language generation.
\newblock \emph{ACM computing surveys}, 55\penalty0 (12):\penalty0 1--38, 2023.

\bibitem[Leng et~al.(2024)Leng, Zhang, Chen, Li, Lu, Miao, and Bing]{vcd}
Sicong Leng, Hang Zhang, Guanzheng Chen, Xin Li, Shijian Lu, Chunyan Miao, and Lidong Bing.
\newblock Mitigating object hallucinations in large vision-language models through visual contrastive decoding.
\newblock In \emph{CVPR}, pages 13872--13882, 2024.

\bibitem[Li et~al.(2023)Li, Du, Zhou, Wang, Zhao, and Wen]{POPE}
Yifan Li, Yifan Du, Kun Zhou, Jinpeng Wang, Wayne~Xin Zhao, and Ji-Rong Wen.
\newblock Evaluating object hallucination in large vision-language models.
\newblock \emph{arXiv preprint arXiv:2305.10355}, 2023.

\bibitem[Liang et~al.(2021)Liang, Glossner, Wang, Shi, and Zhang]{liang2021pruning}
Tailin Liang, John Glossner, Lei Wang, Shaobo Shi, and Xiaotong Zhang.
\newblock Pruning and quantization for deep neural network acceleration: A survey.
\newblock \emph{Neurocomputing}, 461:\penalty0 370--403, 2021.

\bibitem[Lin et~al.(2014)Lin, Maire, Belongie, Hays, Perona, Ramanan, Doll{\'a}r, and Zitnick]{coco}
Tsung-Yi Lin, Michael Maire, Serge Belongie, James Hays, Pietro Perona, Deva Ramanan, Piotr Doll{\'a}r, and C~Lawrence Zitnick.
\newblock Microsoft coco: Common objects in context.
\newblock In \emph{ECCV}, pages 740--755. Springer, 2014.

\bibitem[Liu et~al.(2024{\natexlab{a}})Liu, Li, Li, and Lee]{llava_1.5}
Haotian Liu, Chunyuan Li, Yuheng Li, and Yong~Jae Lee.
\newblock Improved baselines with visual instruction tuning.
\newblock In \emph{CVPR}, pages 26296--26306, 2024{\natexlab{a}}.

\bibitem[Liu et~al.(2024{\natexlab{b}})Liu, Xue, Chen, Chen, Zhao, Wang, Hou, Li, and Peng]{liu2024survey}
Hanchao Liu, Wenyuan Xue, Yifei Chen, Dapeng Chen, Xiutian Zhao, Ke Wang, Liping Hou, Rongjun Li, and Wei Peng.
\newblock A survey on hallucination in large vision-language models.
\newblock \emph{arXiv preprint arXiv:2402.00253}, 2024{\natexlab{b}}.

\bibitem[Michel et~al.(2019)Michel, Levy, and Neubig]{michel2019sixteen}
Paul Michel, Omer Levy, and Graham Neubig.
\newblock Are sixteen heads really better than one?
\newblock \emph{NeurIPS}, 32, 2019.

\bibitem[Oquab et~al.(2023)Oquab, Darcet, Moutakanni, Vo, Szafraniec, Khalidov, Fernandez, Haziza, Massa, El-Nouby, et~al.]{dinov2}
Maxime Oquab, Timoth{\'e}e Darcet, Th{\'e}o Moutakanni, Huy Vo, Marc Szafraniec, Vasil Khalidov, Pierre Fernandez, Daniel Haziza, Francisco Massa, Alaaeldin El-Nouby, et~al.
\newblock Dinov2: Learning robust visual features without supervision.
\newblock \emph{arXiv preprint arXiv:2304.07193}, 2023.

\bibitem[Rohrbach et~al.(2018)Rohrbach, Hendricks, Burns, Darrell, and Saenko]{chair}
Anna Rohrbach, Lisa~Anne Hendricks, Kaylee Burns, Trevor Darrell, and Kate Saenko.
\newblock Object hallucination in image captioning.
\newblock \emph{arXiv preprint arXiv:1809.02156}, 2018.

\bibitem[Shu et~al.(2025)Shu, Zhao, Hu, Liu, Payani, Cheng, and Du]{shu2025large}
Dong Shu, Haiyan Zhao, Jingyu Hu, Weiru Liu, Ali Payani, Lu Cheng, and Mengnan Du.
\newblock Large vision-language model alignment and misalignment: A survey through the lens of explainability.
\newblock \emph{arXiv preprint arXiv:2501.01346}, 2025.

\bibitem[Sim{\'e}oni et~al.(2021)Sim{\'e}oni, Puy, Vo, Roburin, Gidaris, Bursuc, P{\'e}rez, Marlet, and Ponce]{simeoni2021localizing}
Oriane Sim{\'e}oni, Gilles Puy, Huy~V Vo, Simon Roburin, Spyros Gidaris, Andrei Bursuc, Patrick P{\'e}rez, Renaud Marlet, and Jean Ponce.
\newblock Localizing objects with self-supervised transformers and no labels.
\newblock \emph{arXiv preprint arXiv:2109.14279}, 2021.

\bibitem[Sim{\'e}oni et~al.(2025)Sim{\'e}oni, Vo, Seitzer, Baldassarre, Oquab, Jose, Khalidov, Szafraniec, Yi, Ramamonjisoa, et~al.]{dinov3}
Oriane Sim{\'e}oni, Huy~V Vo, Maximilian Seitzer, Federico Baldassarre, Maxime Oquab, Cijo Jose, Vasil Khalidov, Marc Szafraniec, Seungeun Yi, Micha{\"e}l Ramamonjisoa, et~al.
\newblock Dinov3.
\newblock \emph{arXiv preprint arXiv:2508.10104}, 2025.

\bibitem[Vo et~al.(2020)Vo, P{\'e}rez, and Ponce]{coco20k}
Huy~V. Vo, Patrick P{\'e}rez, and Jean Ponce.
\newblock Toward unsupervised, multi-object discovery in large-scale image collections.
\newblock In \emph{ECCV}, pages 779--795, Cham, 2020. Springer International Publishing.

\bibitem[Voita et~al.(2019)Voita, Talbot, Moiseev, Sennrich, and Titov]{voita2019analyzing}
Elena Voita, David Talbot, Fedor Moiseev, Rico Sennrich, and Ivan Titov.
\newblock Analyzing multi-head self-attention: Specialized heads do the heavy lifting, the rest can be pruned.
\newblock \emph{arXiv preprint arXiv:1905.09418}, 2019.

\bibitem[Wang et~al.(2023{\natexlab{a}})Wang, Zhou, Xu, Shi, Zhao, Xu, Ye, Yan, Zhang, Zhu, et~al.]{wang2023evaluation}
Junyang Wang, Yiyang Zhou, Guohai Xu, Pengcheng Shi, Chenlin Zhao, Haiyang Xu, Qinghao Ye, Ming Yan, Ji Zhang, Jihua Zhu, et~al.
\newblock Evaluation and analysis of hallucination in large vision-language models.
\newblock \emph{arXiv preprint arXiv:2308.15126}, 2023{\natexlab{a}}.

\bibitem[Wang et~al.(2023{\natexlab{b}})Wang, Shen, Yuan, Du, Li, Hu, Crowley, and Vaufreydaz]{tokencut}
Yangtao Wang, Xi Shen, Yuan Yuan, Yuming Du, Maomao Li, Shell~Xu Hu, James~L Crowley, and Dominique Vaufreydaz.
\newblock Tokencut: Segmenting objects in images and videos with self-supervised transformer and normalized cut.
\newblock \emph{IEEE TPAMI}, 45\penalty0 (12):\penalty0 15790--15801, 2023{\natexlab{b}}.

\bibitem[Wang et~al.(2019)Wang, Wohlwend, and Lei]{wang2019structured}
Ziheng Wang, Jeremy Wohlwend, and Tao Lei.
\newblock Structured pruning of large language models.
\newblock \emph{arXiv preprint arXiv:1910.04732}, 2019.

\bibitem[Woo et~al.(2024)Woo, Jang, Kim, Choi, and Kim]{ritual}
Sangmin Woo, Jaehyuk Jang, Donguk Kim, Yubin Choi, and Changick Kim.
\newblock Ritual: Random image transformations as a universal anti-hallucination lever in large vision language models.
\newblock \emph{arXiv preprint arXiv:2405.17821}, 2024.

\bibitem[Yin et~al.(2024{\natexlab{a}})Yin, Fu, Zhao, Li, Sun, Xu, and Chen]{mme_data}
Shukang Yin, Chaoyou Fu, Sirui Zhao, Ke Li, Xing Sun, Tong Xu, and Enhong Chen.
\newblock A survey on multimodal large language models.
\newblock \emph{National Science Review}, 11\penalty0 (12):\penalty0 nwae403, 2024{\natexlab{a}}.

\bibitem[Yin et~al.(2024{\natexlab{b}})Yin, Fu, Zhao, Xu, Wang, Sui, Shen, Li, Sun, and Chen]{woodpecker}
Shukang Yin, Chaoyou Fu, Sirui Zhao, Tong Xu, Hao Wang, Dianbo Sui, Yunhang Shen, Ke Li, Xing Sun, and Enhong Chen.
\newblock Woodpecker: Hallucination correction for multimodal large language models.
\newblock \emph{Science China Information Sciences}, 67\penalty0 (12):\penalty0 220105, 2024{\natexlab{b}}.

\bibitem[Zhang et~al.(2025)Zhang, Wan, Kan, Ma, Stepputtis, Ramanan, Salakhutdinov, Morency, Sycara, and Xie]{degf}
Ce Zhang, Zifu Wan, Zhehan Kan, Martin~Q Ma, Simon Stepputtis, Deva Ramanan, Russ Salakhutdinov, Louis-Philippe Morency, Katia Sycara, and Yaqi Xie.
\newblock Self-correcting decoding with generative feedback for mitigating hallucinations in large vision-language models.
\newblock \emph{arXiv preprint arXiv:2502.06130}, 2025.

\bibitem[Zhao et~al.(2024)Zhao, Deng, Zhang, and Gu]{marine}
Linxi Zhao, Yihe Deng, Weitong Zhang, and Quanquan Gu.
\newblock Mitigating object hallucination in large vision-language models via image-grounded guidance.
\newblock \emph{arXiv preprint arXiv:2402.08680}, 2024.

\bibitem[Zhou et~al.(2022)Zhou, Yu, Xie, Xiao, Anandkumar, Feng, and Alvarez]{zhou2022understanding}
Daquan Zhou, Zhiding Yu, Enze Xie, Chaowei Xiao, Animashree Anandkumar, Jiashi Feng, and Jose~M Alvarez.
\newblock Understanding the robustness in vision transformers.
\newblock In \emph{ICML}, pages 27378--27394. PMLR, 2022.

\end{thebibliography}
}

\clearpage
\setcounter{page}{1}
\maketitlesupplementary

This supplementary material provides additional experimental details and ablation studies to support our work. We show: (1) Evaluation of the optimal number of clusters K for our head selection algorithm using the Davies-Bouldin Score, (2) Ablation study on the guidance weight $\alpha$ in our MLLM hallucination mitigation method, (3) Analysis of the temperature parameter $\tau$ from Equation \ref{eq:self_sim} and its impact on object-centric signal quality, and a visualization demonstrating how visual patterns evolve across different temperature values, (4) Visualization of individual head behaviors in the final layer, (5) Ablation study of ensemble weight $W_q, W_k, W_v$ in eq. \ref{eq:ensemble_sim}, (6) Computational efficiency analysis for MLLM hallucination mitigation, (7) Analysis of object-centric heads across ViT architecture and reconstruction-based self-supervised models, (8) Evaluation Metrics and  Datasets used. (9) Qualitative comparison of MLLM-generated captions. 

\section{Optimal Number of Clusters K}
\label{supp_sec:optimal_k}
A critical hyperparameter in our Object-DINO algorithm is the number of clusters K used in the k-means clustering step (Algorithm 1, line 13). To determine the optimal value, we evaluate multiple candidate values of K using the Davies-Bouldin (DB) Score, a standard metric for assessing clustering quality.

\paragraph{Davies-Bouldin Score}

The Davies-Bouldin Score measures the average similarity between each cluster and its most similar cluster, indicating cluster separation and compactness. The equation is given by:
\begin{equation}
\text{DB}(K) = \frac{1}{K} \sum_{i=1}^{K} \max_{j \neq i} \left( \frac{s_i + s_j}{d_{ij}} \right)
\end{equation}
where, $s_i$ is the average distance between each point in cluster $i$ and the cluster centroid $d_{ij}$ is the distance between cluster centroids $c_i$ and $c_j$

\textbf{Experimental Setup.} We evaluate K $\in$ {3, 4, 5, 6, 7, 8, 9, 10} on 500 randomly sampled images from the COCO dataset using DINO-V3 features. For each value of K, we compute the DB Score.

\begin{figure}
    \centering
        \includegraphics[width=\linewidth]{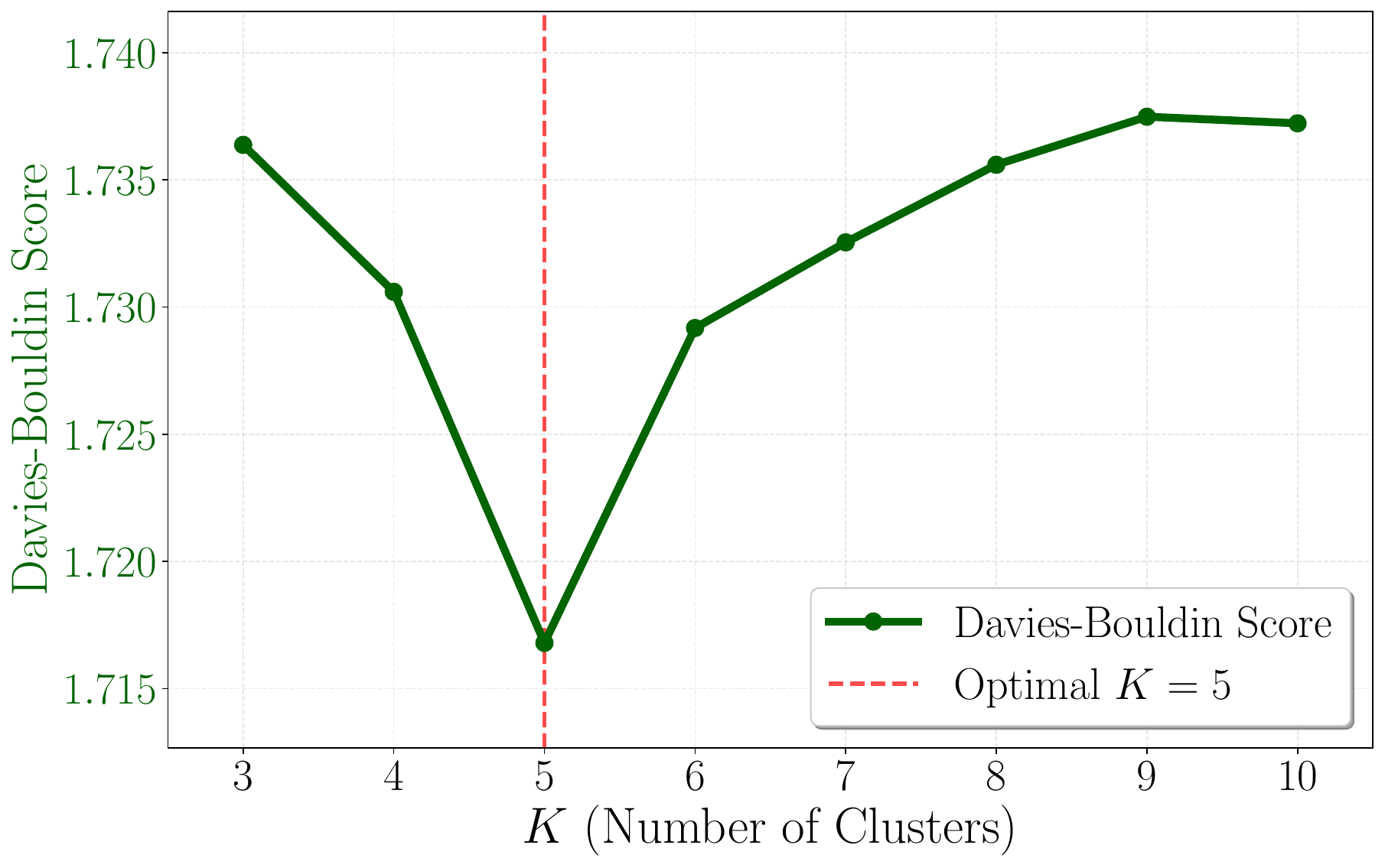}
    \caption{Evaluation of optimal number of clusters $K$ using Davies-Bouldin Score (lower is better). We select $K=5$ for all our experiments.}
    \label{fig:kmeans}
\end{figure}

\textbf{Results} As shown in Fig. \ref{fig:kmeans}, the DB Score remains relatively stable across the evaluated range, varying between 1.717 and 1.737. The minimum score occurs at K = 5, suggesting this configuration achieves the best balance between cluster separation and compactness.  Based on this, we use K = 5 for all experiments in the main paper.

\section{Ablation on Guidance Weight for MLLM Hallucination Mitigation}
\label{supp_sec:alpha_ablation}

\begin{figure}
    \centering
    \includegraphics[width=\linewidth]{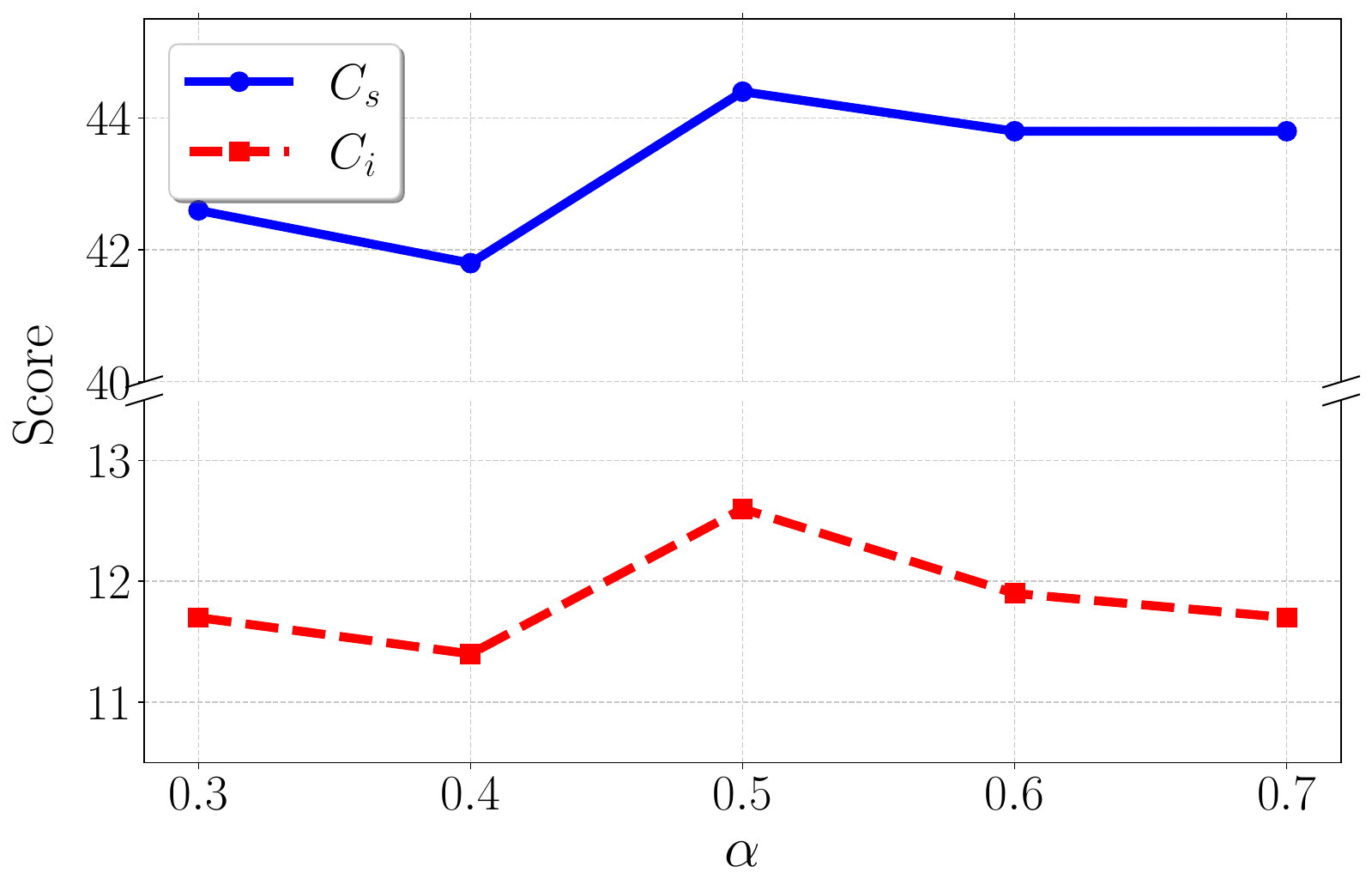}
    \caption{Guidance weight ablation for MLLM hallucination mitigation. We observe that the $C_s$ and $C_i$ scores are lowest for $\alpha$ = 0.4, making it our optimal choice}
    \label{fig:alpha_ablation}
\end{figure}

Our MLLM hallucination mitigation method (Sec. \ref{sec:mllm_hall}) introduces a guidance weight $\alpha$ that controls the strength of the visual grounding signal (Eq. \ref{eq:mllm_combine}):
\begin{equation} 
 L =\alpha \ \text{Logits}(y|T_u, R, u) + (1-\alpha)\text{Logits}(y|T_v, R, v) 
\end{equation}
 When we set $\alpha$ = 0, it is equivalent to the original MLLM output with regular decoding, while a high alpha ensures grounding information from the Object-DINO map. We perform an ablation by varying $\alpha$ across a range ([0.3, 0.4, 0.5, 0.6, 0.7]) and evaluate performance on the CHAIR benchmark. Based on the Fig. \ref{fig:alpha_ablation}, we select $\alpha$ = 0.4 for our experiments as it demonstrates the lowest score for both $C_s$ and $C_i$.

\section{Ablation on Temperature Parameter}
\label{sec_supp:ablate_temp}
The temperature parameter $\tau$ in Eq. \ref{eq:self_sim} controls the sharpness of the patch self-similarity matrices ($A_q, A_k, A_v$). It scales the dot-product operation before the softmax:
\begin{equation}
A_r^{\ell,h} = \text{softmax}\left( \frac{ \tilde{r}^{\ell,h} \cdot (\tilde{r}^{\ell,h})^\top }{ \tau } \right)
\end{equation}

\begin{figure}
    \centering
    \includegraphics[width=\linewidth]{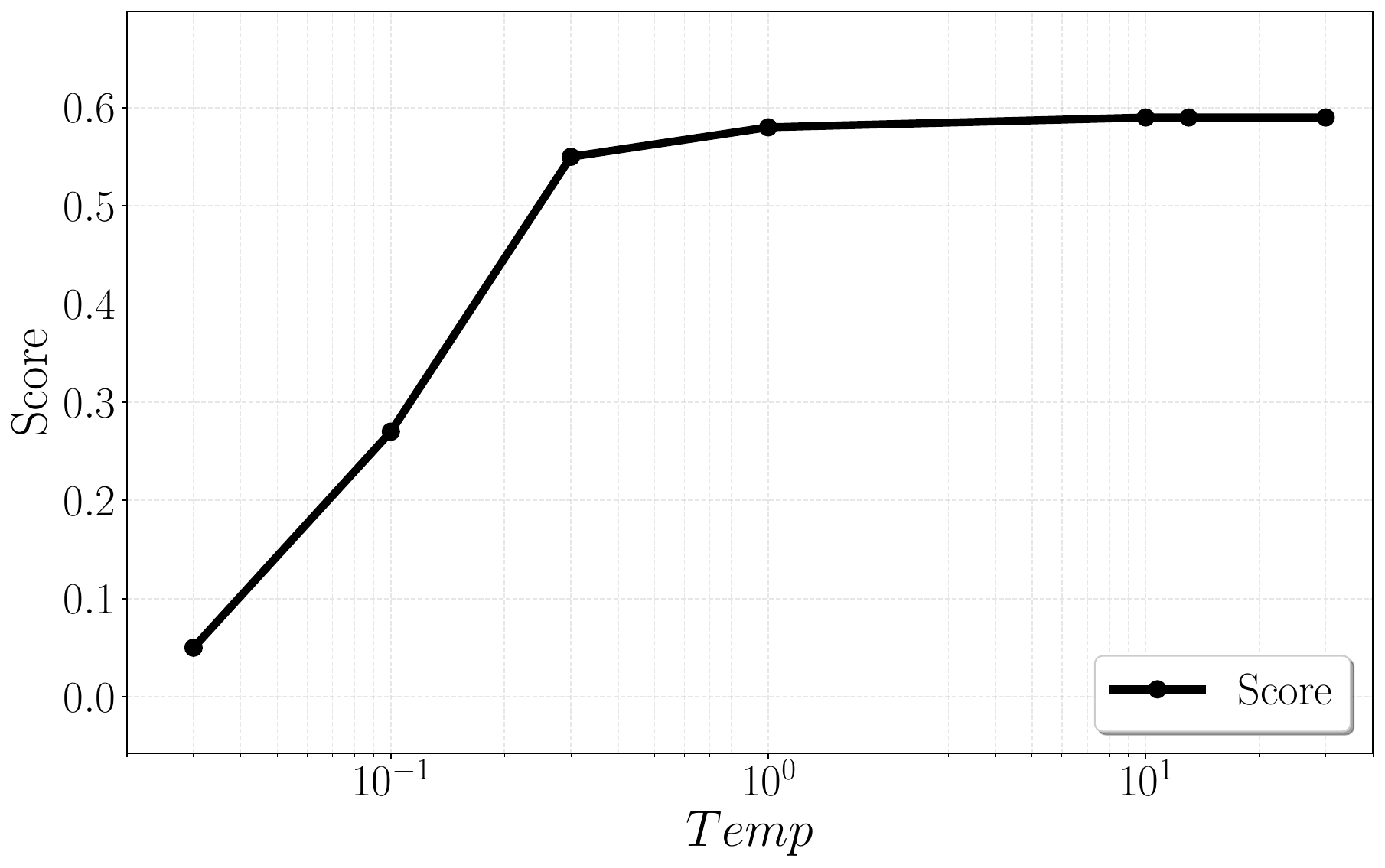}
    \caption{\textbf{Tau ablation.} We show the optimal choice of $\tau$ for object centric signal is $\in (10,100)$}
    \label{fig:temp_ablation}
\end{figure}

\begin{figure*}

    \centering
    \begin{tabular}{@{\hskip 0pt}c@{\hskip 2pt}c@{\hskip 2pt}c@{\hskip 2pt}c@{\hskip 0pt}}
        $\tau = 0.03$ & $\tau = 0.1$ & $\tau =0.3$ & $\tau = 1$ \\
        \includegraphics[width=0.20\textwidth]{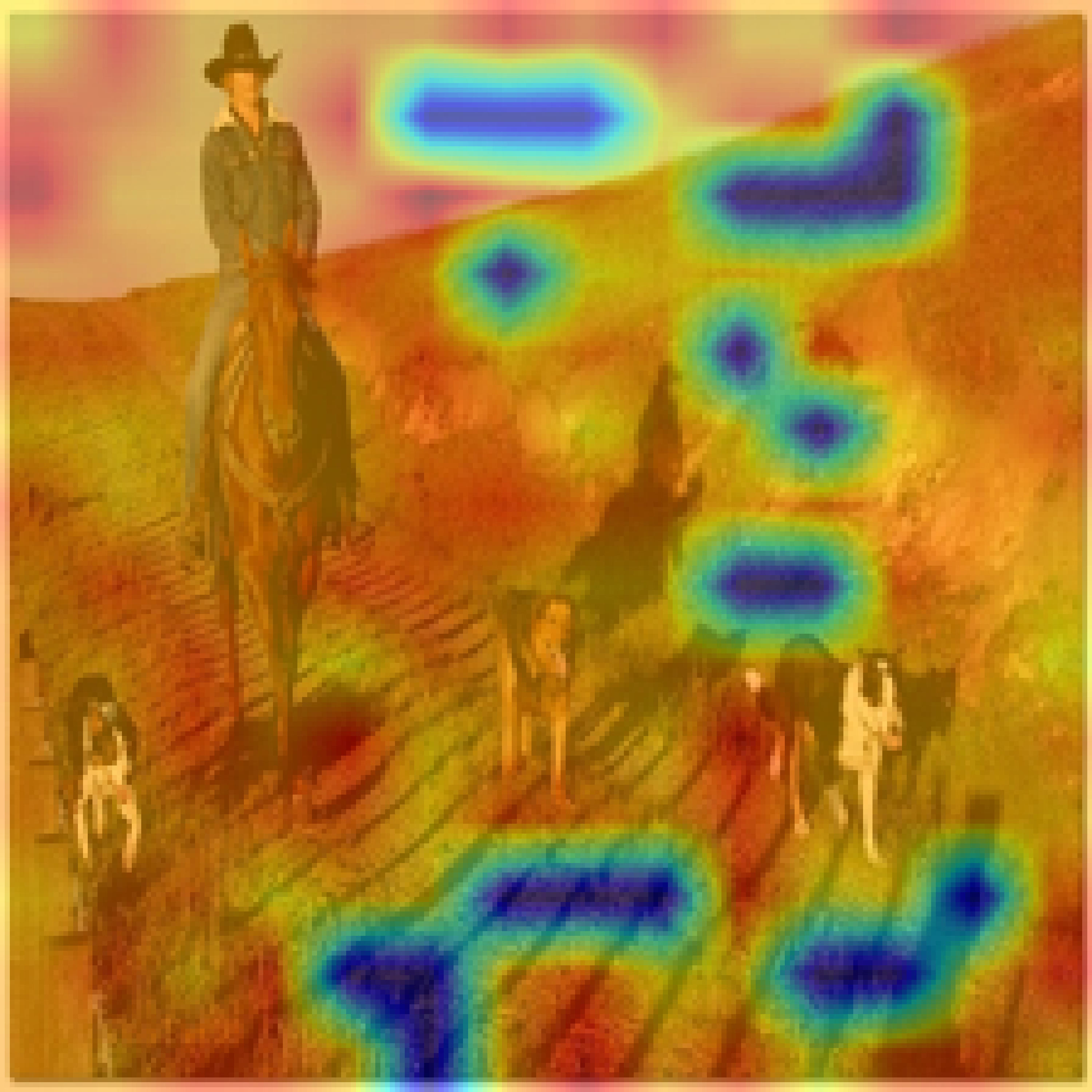} &
        \includegraphics[width=0.20\textwidth]{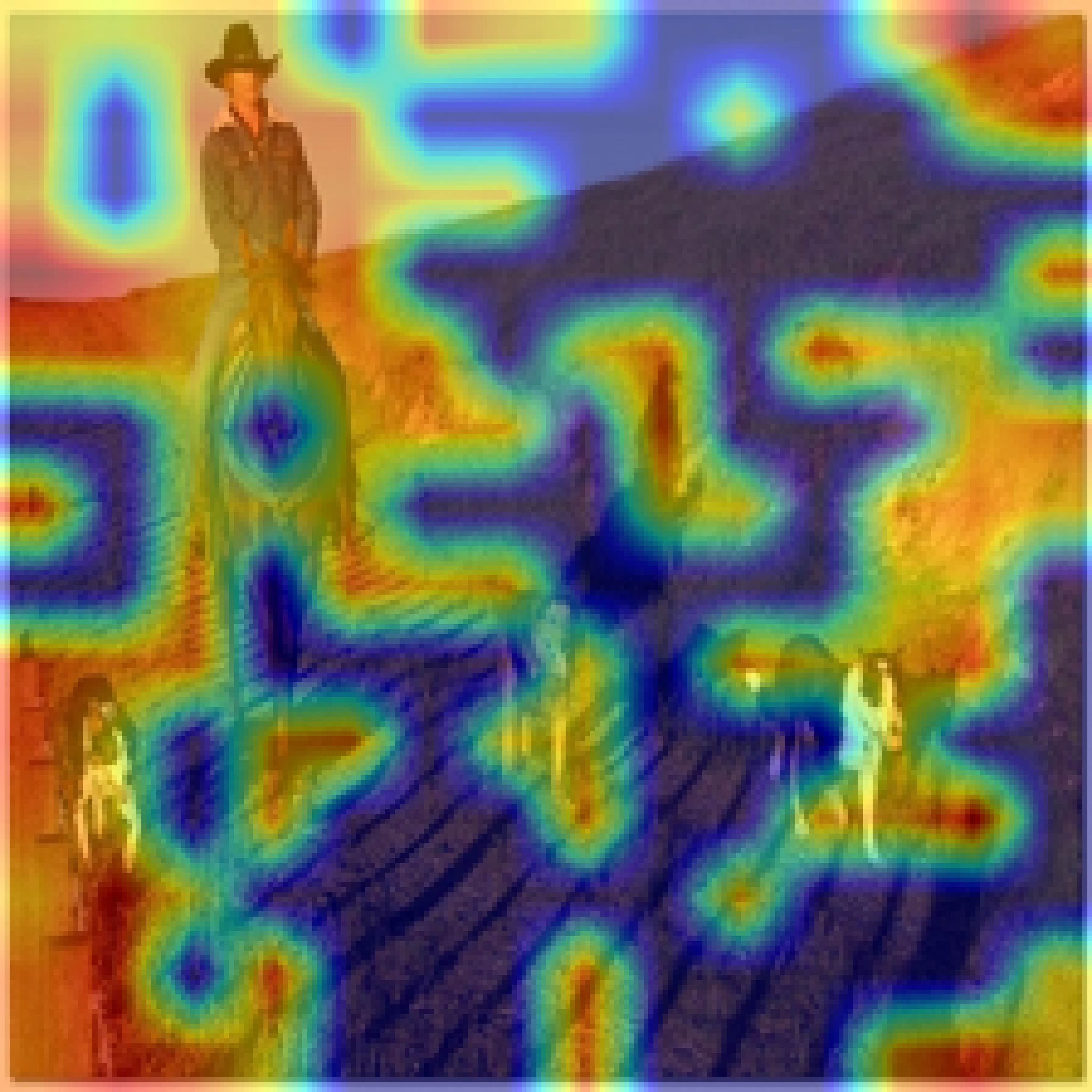} &
        \includegraphics[width=0.20\textwidth]{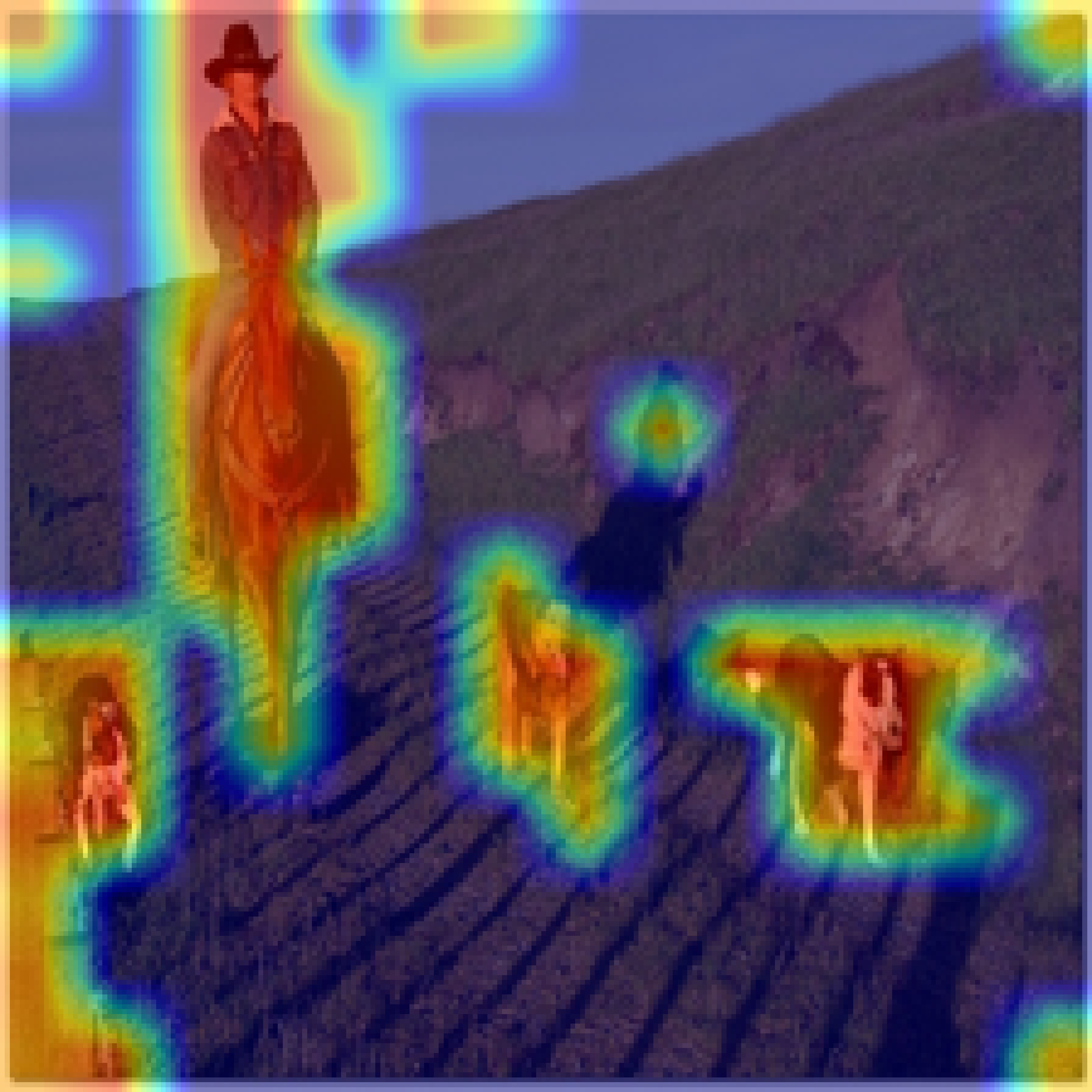} &
        \includegraphics[width=0.20\textwidth]{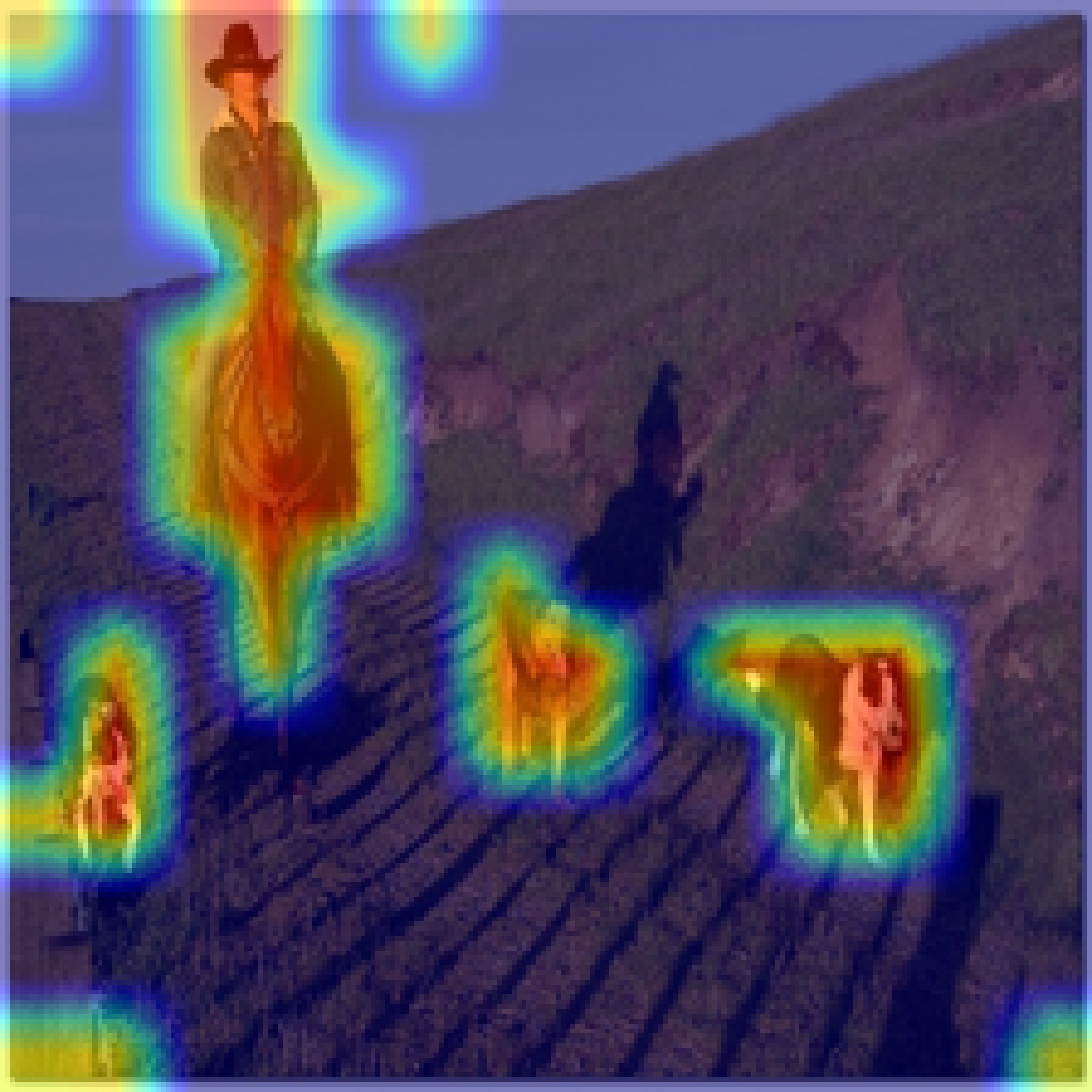} \\

        $\tau = 3$ & $\tau =10$ & $\tau = 30$ & $\tau = 100$ \\
        \includegraphics[width=0.20\textwidth]{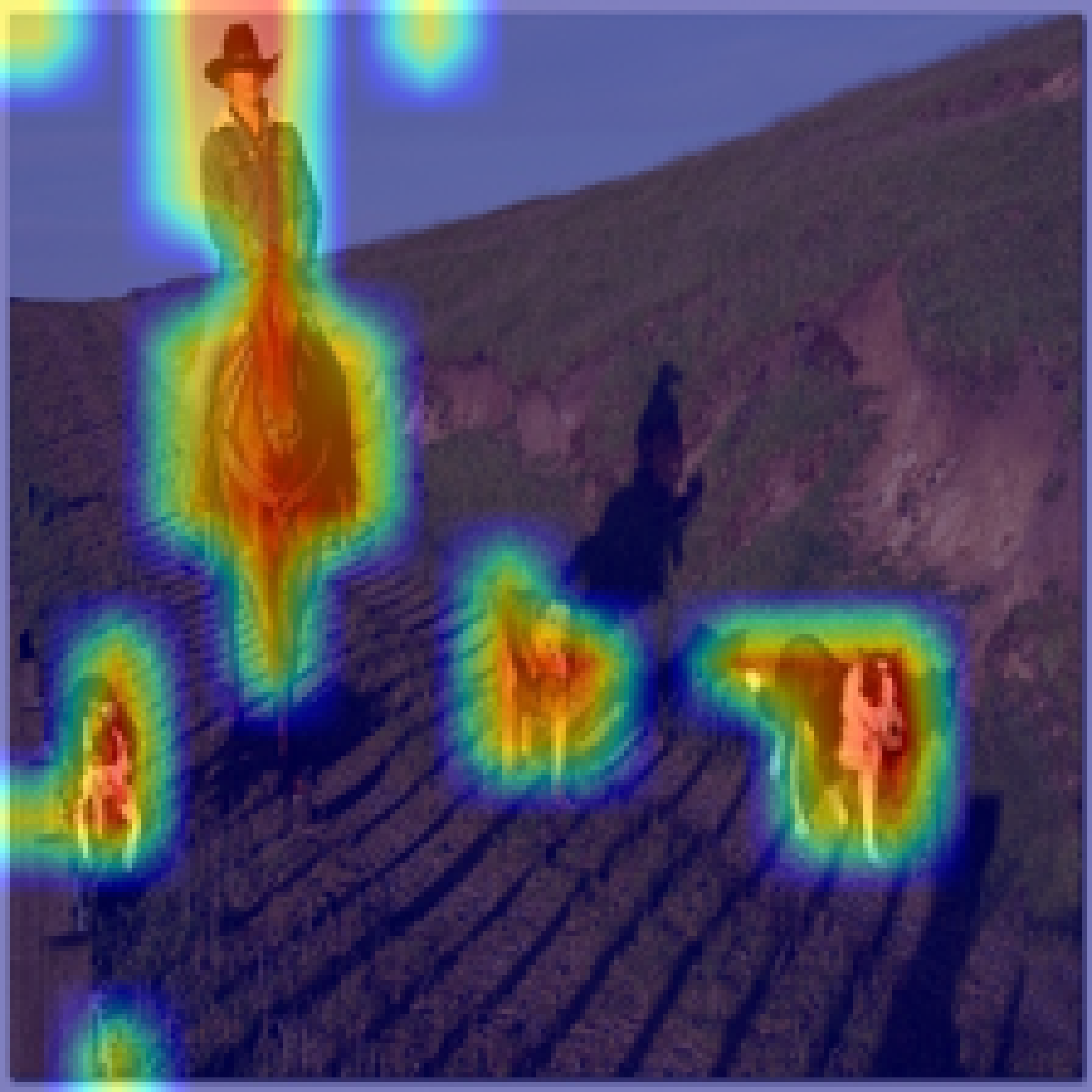}  &
        \includegraphics[width=0.20\textwidth]{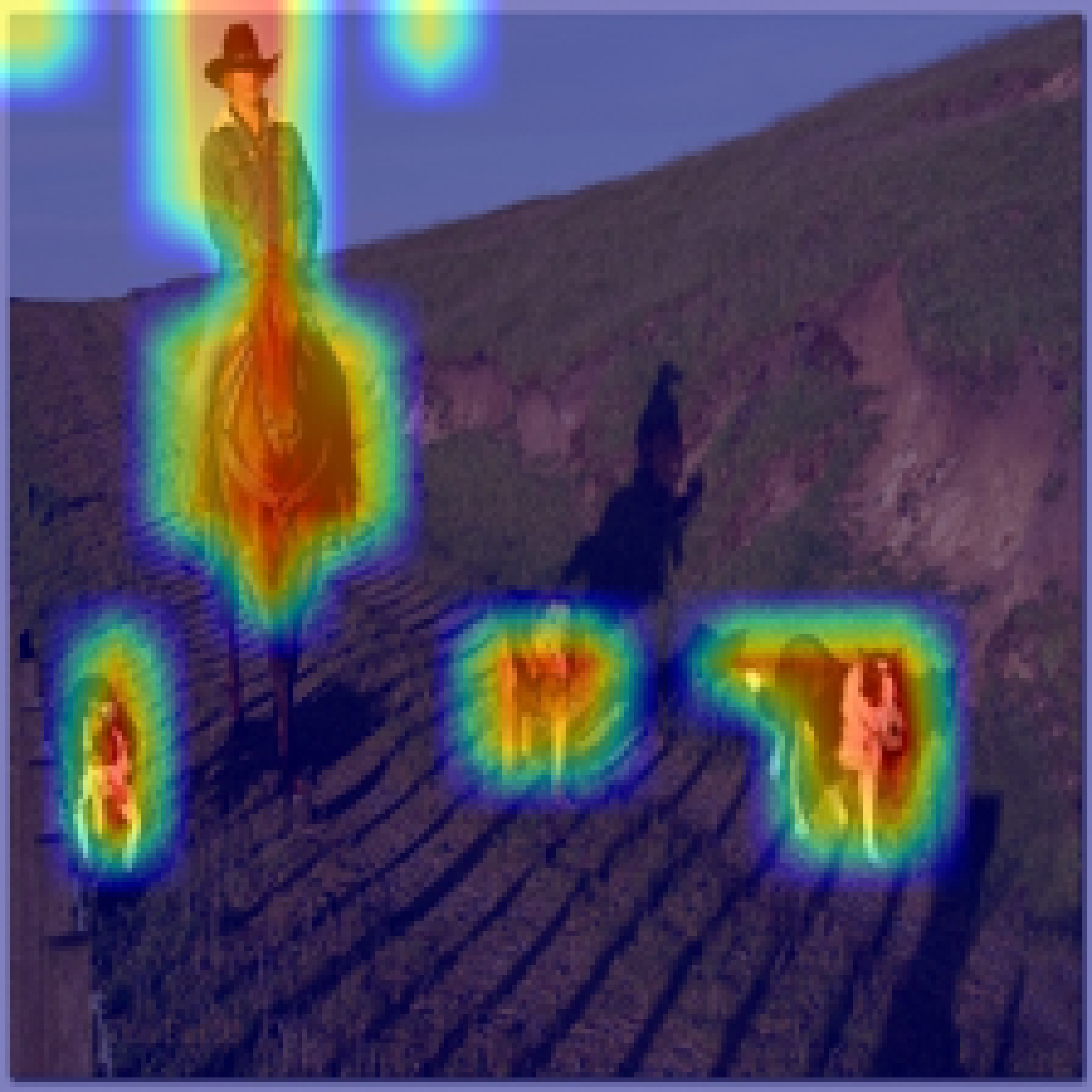} &
        \includegraphics[width=0.20\textwidth]{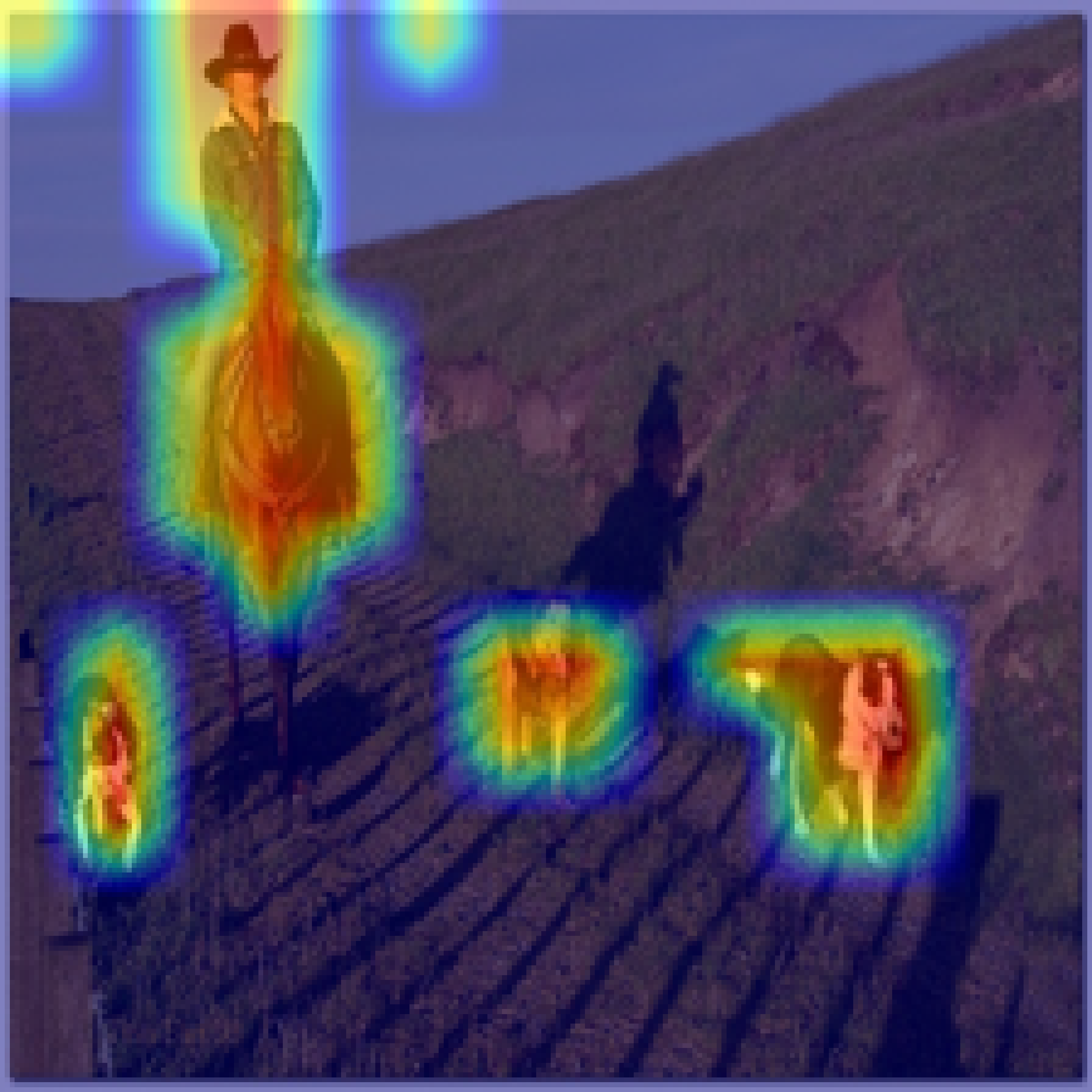} &
        \includegraphics[width=0.20\textwidth]{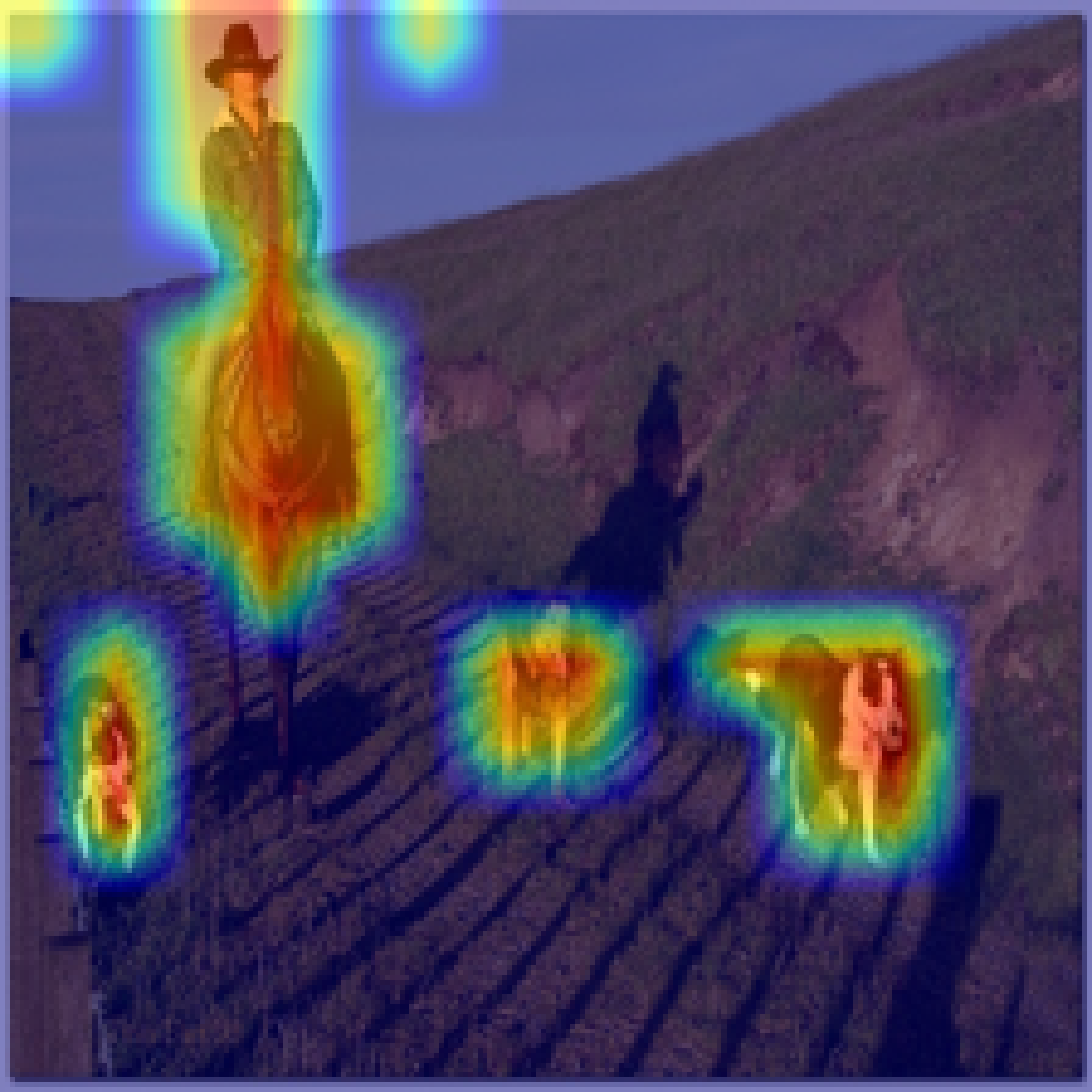} \\
      
    \end{tabular}
    \captionof{figure}{Evolution of localizations with temperature. For visualization, we invert the maps to show bright colors for objects.}
    \label{fig:temp_evol}
\end{figure*}

We investigate how $\tau$ affects the quality of object-centric signals. We evaluate $\tau \in$ {0.03, 0.1, 0.3, 1, 3, 10, 30, 100} on a subset of COCO images. Fig. \ref{fig:temp_ablation} shows the variation of $\tau$ and how it impacts the object-centric score. Based on this, we identify that selecting a $\tau \in (10,100)$ leads to optimal object-centric signals. We further show the evolution of $\tau$ in Fig. \ref{fig:temp_evol}.

\section{Final Layer Head Visualization}
\label{supp_sec:head_viz}
\begin{figure*}

    \centering
    \begin{tabular}{@{\hskip 0pt}c@{\hskip 2pt}c@{\hskip 2pt}c@{\hskip 2pt}c@{\hskip 0pt}}
        Head 0 & Head 1 & Head 2 & Head 3 \\
        \includegraphics[width=0.20\textwidth]{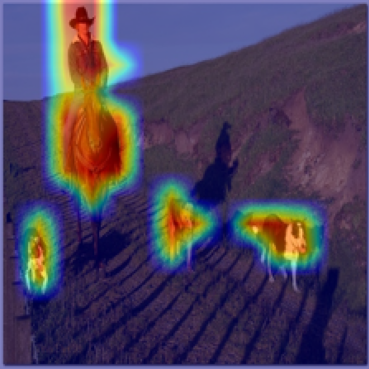} &
        \includegraphics[width=0.20\textwidth]{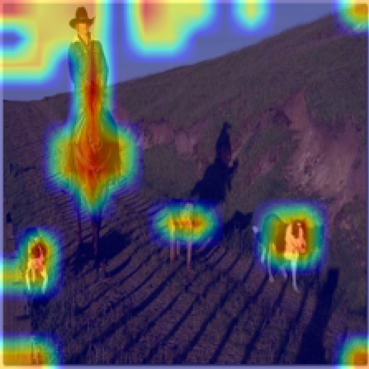} &
        \includegraphics[width=0.20\textwidth]{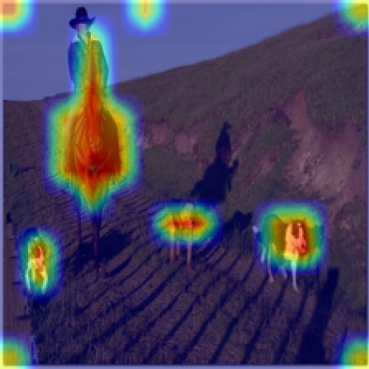} &
        \includegraphics[width=0.20\textwidth]{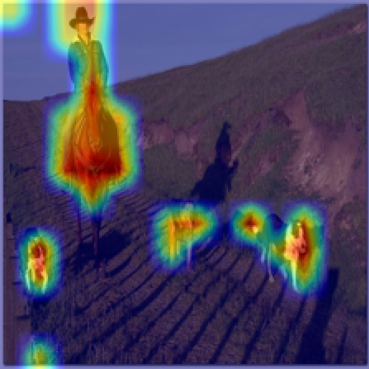} \\

        Head 4 & Head 5 & Head 6 & Head 7 \\
        \includegraphics[width=0.20\textwidth]{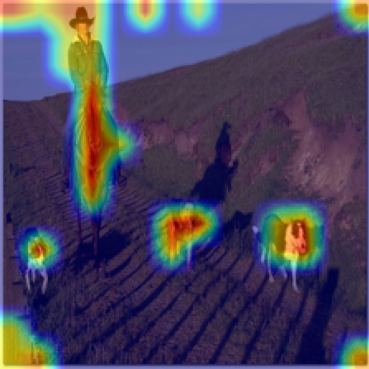}  &
        \includegraphics[width=0.20\textwidth]{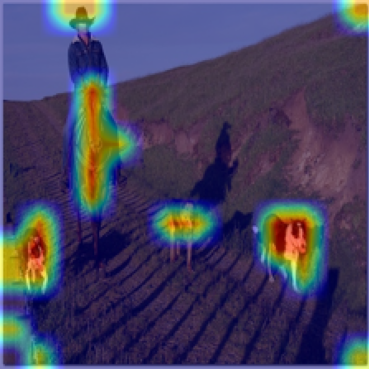} &
        \includegraphics[width=0.20\textwidth]{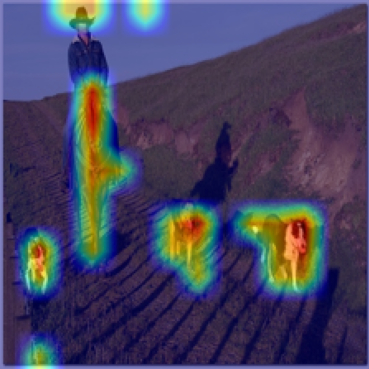} &
        \includegraphics[width=0.20\textwidth]{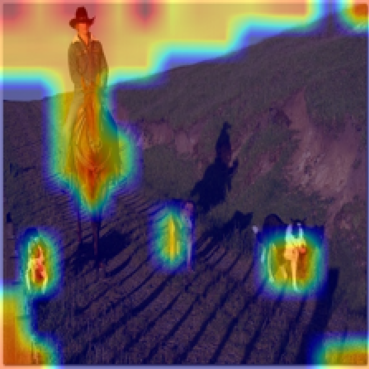} \\

        Head 8 & Head 9 & Head 10 & Head 11 \\
        \includegraphics[width=0.20\textwidth]{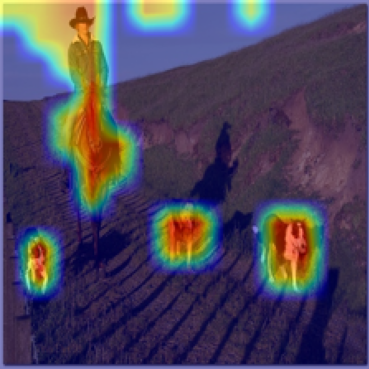}  &
        \includegraphics[width=0.20\textwidth]{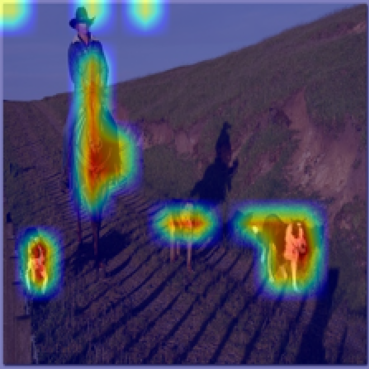} &
        \includegraphics[width=0.20\textwidth]{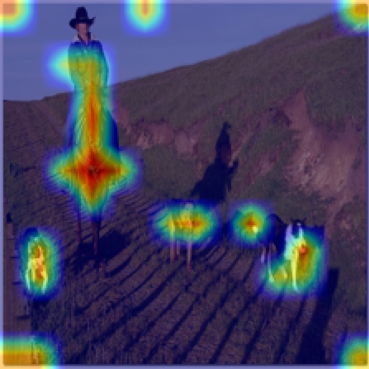} &
        \includegraphics[width=0.20\textwidth]{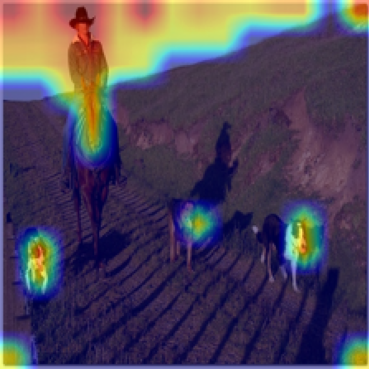}
    \end{tabular}
    \captionof{figure}{\textbf{Final Layer Head Visualization.} Ensemble similarity maps $A_{ens}$ for all 12 heads in Layer 11 of DINO-V3. For visualization, we invert the maps to show bright colors for objects. }
    \label{fig:final_layer_head_visualization}
\end{figure*}

To illustrate the diversity of attention head specializations and support our claim that not all final-layer heads are object-centric, in fig. \ref{fig:final_layer_head_visualization} we visualize the ensemble similarity maps ($A_{ens}$) for all 12 attention heads in the final layer (Layer 11) of DINO-V3. These plots are consistent with the analysis shown in Fig. \ref{fig:analysis_figs}, where heads 5, 6, 7, and 11 are not selected as object-centric because head 5 and head 6 miss regions of the man, and head 7 and head 11 include a large portion of the background.

\section{Ablation of Ensemble Weights}
\label{supp_sec:weight_ablation}
In this section, we ablate the ensemble weights ($W_q, W_k, W_v$) from Eq. \ref{eq:ensemble_sim}. The results in Table 6. shows CorLoc performance on VOC2007 with different weight configurations. 
We observe that heavily weighting the Key component $W_k = 0.7$ yields a CorLoc of 17.9. Performance progressively increases as the weights become more balanced, peaking at 19.8 CorLoc with a uniform average ($W_q = W_k = W_v$ = 0.33). This confirms that all three components contribute valuable and complementary object signals. We use this average for $A_{ens}$ in all experiments.

\begin{table}[tp]
\captionof{table}{\textbf{Ablation of Ensemble Weights.} We report CorLoc for different weight configurations from Eq. \ref{eq:ensemble_sim}. Performance peaks at 19.8 with a uniform average ($W_q = W_k = W_v$ = 0.33), confirming that all three components contribute valuable and complementary object signals. We therefore use this uniform average for $A_{ens}$ in all experiments.}

\label{tab:w_ablation}
\centering
\resizebox{0.65\columnwidth}{!}{
\begin{tabular}{cccc}
\toprule
\textbf{$W_q$} & \textbf{$W_v$} & \textbf{$W_k$} & \textbf{CorLoc} \\
\midrule
\rowcolor{lightgray!45} 0.1 & 0.2 & 0.7 & 17.9 \\
0.2 & 0.3 & 0.5 & 19.1 \\
\rowcolor{lightgray!45} 0.3 & 0.3 & 0.4 & 19.4 \\
0.33 & 0.33 & 0.33 & 19.8 \\
\bottomrule
\end{tabular}

}
\end{table}

\section{Computational Efficiency Analysis for MLLM Hallucination Mitigation}
\label{Supp_sec:MLLM_efficiency}
In Table. \ref{tab:latency} we compare the efficiency and performance of our method against prior works for mitigating object hallucination on the LLaVA-1.5 model. Our approach, which leverages Object-DINO as a training-free visual grounding mechanism, achieves a state-of-the-art CHAIR$_{\text{S}}$ score of 41.6. This significantly outperforms all competing methods, including DeGF (48.8) and HALC (51). Critically, this performance is achieved without the substantial computational overhead seen by other methods. While approaches like OPERA, HALC, and DeGF incur massive latency penalties (ranging from 4.04x to 7.18x the regular baseline) and significant GPU memory increases (1.21x to 1.46x), our method maintains a minimal computational cost. Our latency (7.1s, 2.06x) and peak memory (16,966MB, 1.07x) are only slightly above the baseline and are comparable to the much lower-performing VCD method, demonstrating a vastly superior balance of performance and efficiency

\begin{table}[tp]
\captionof{table}{\textbf{Efficiency comparison.} For each method, we show the inference latency per instance, peak GPU memory and the performance CHAIR$_{\text{S}}$ on the LLaVA-1.5 model with max token 128}
\label{tab:latency}
\resizebox{\columnwidth}{!}{
\begin{tabular}{lcccccc}
\toprule
\textbf{Method} & \textbf{Avg. Latency (s)} & \textbf{GPU Memory (MB)} & \textbf{CHAIR$_{\text{S}}$} ($\downarrow$) \\
\midrule
\rowcolor{lightgray!45} Regular & 3.44 (×1.00) & 15,778 (×1.00) & 55 \\
VCD & 6.91 (×2.01) & 16,634 (×1.05) & 54.4 \\
\rowcolor{lightgray!45}OPERA & 24.70 (×7.18) & 22,706 (×1.44) & 52.6 \\
Woodpecker & 10.68 (×3.10) & 22,199 (×1.41) & 57.6 \\
\rowcolor{lightgray!45}HALC & 22.61 (×6.51) & 23,084 (×1.46) & 51 \\
 DeGF & 13.89 (×4.04) & 19,119 (×1.21) & 48.8 \\
\midrule
\rowcolor{orange!35}Ours & 7.1 (×2.06) & 16,966 (×1.07) & 41.6 \\
\bottomrule
\end{tabular}

}
\end{table}

\begin{figure*}[tp]
    \centering
    \resizebox{0.95\textwidth}{!}{%
        \begin{tabular}{ccc}
            \begin{minipage}{0.32\textwidth}
                \centering
                \includegraphics[width=\linewidth,height=0.75\linewidth,keepaspectratio]{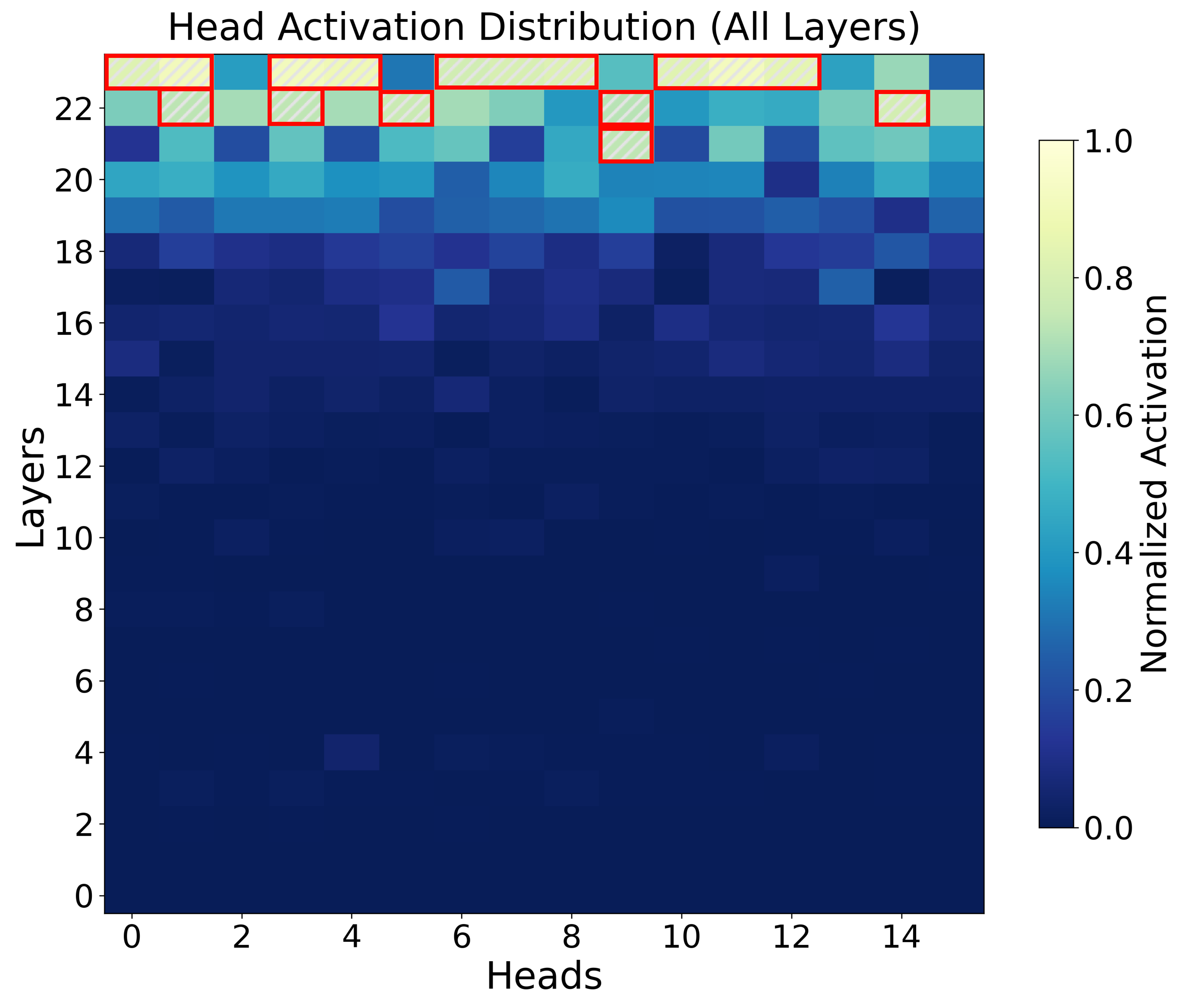}
                \caption*{(a) Head contribution for complete network}
            \end{minipage} &
            \begin{minipage}{0.32\textwidth}
                \centering
                \includegraphics[width=\linewidth,height=0.85\linewidth,keepaspectratio]{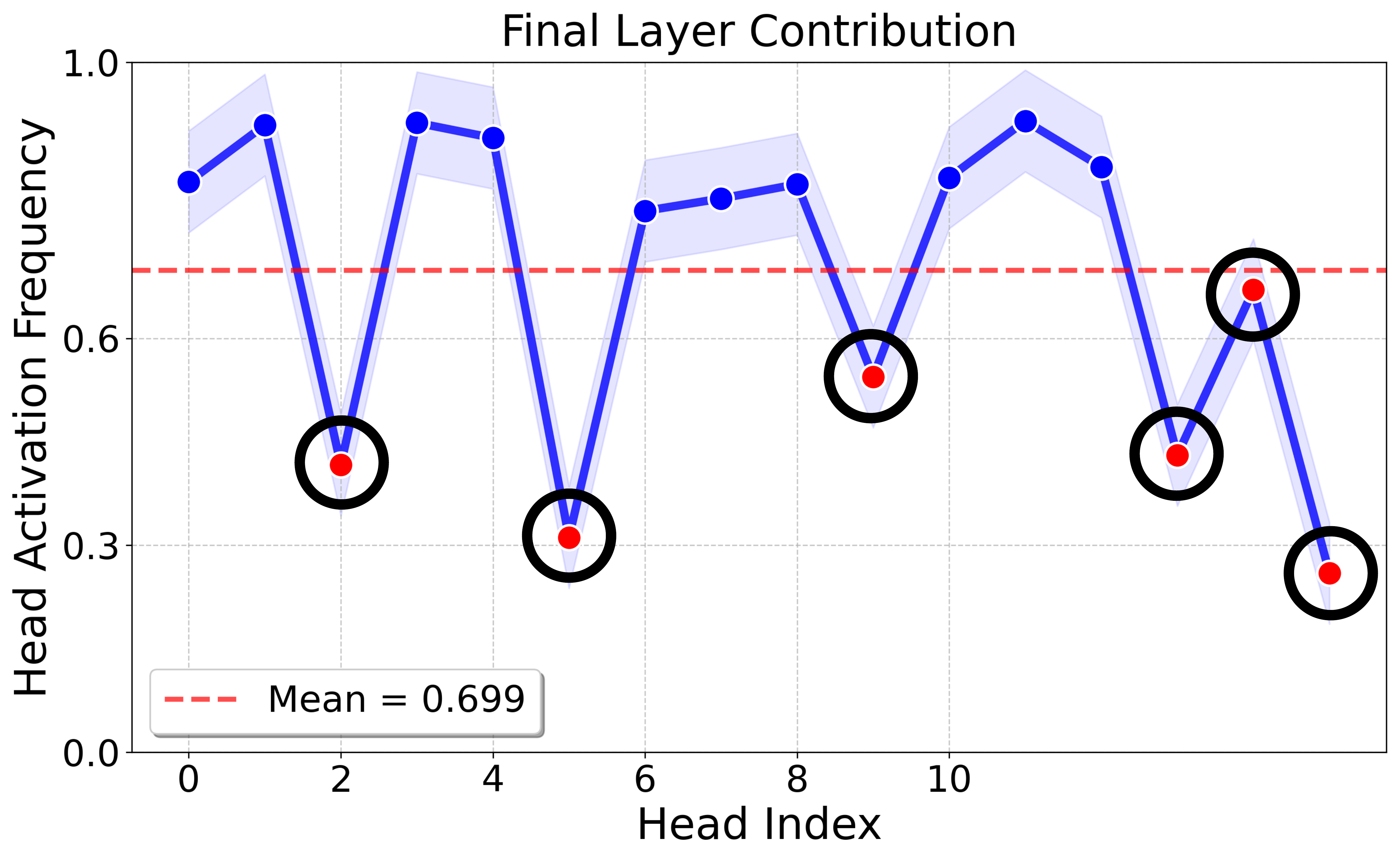}
                \caption*{(b) Heads contribution from final layer.}
            \end{minipage} &
            \begin{minipage}{0.33\textwidth}
                \centering
                \includegraphics[width=0.8\linewidth,height=0.8\linewidth,keepaspectratio]{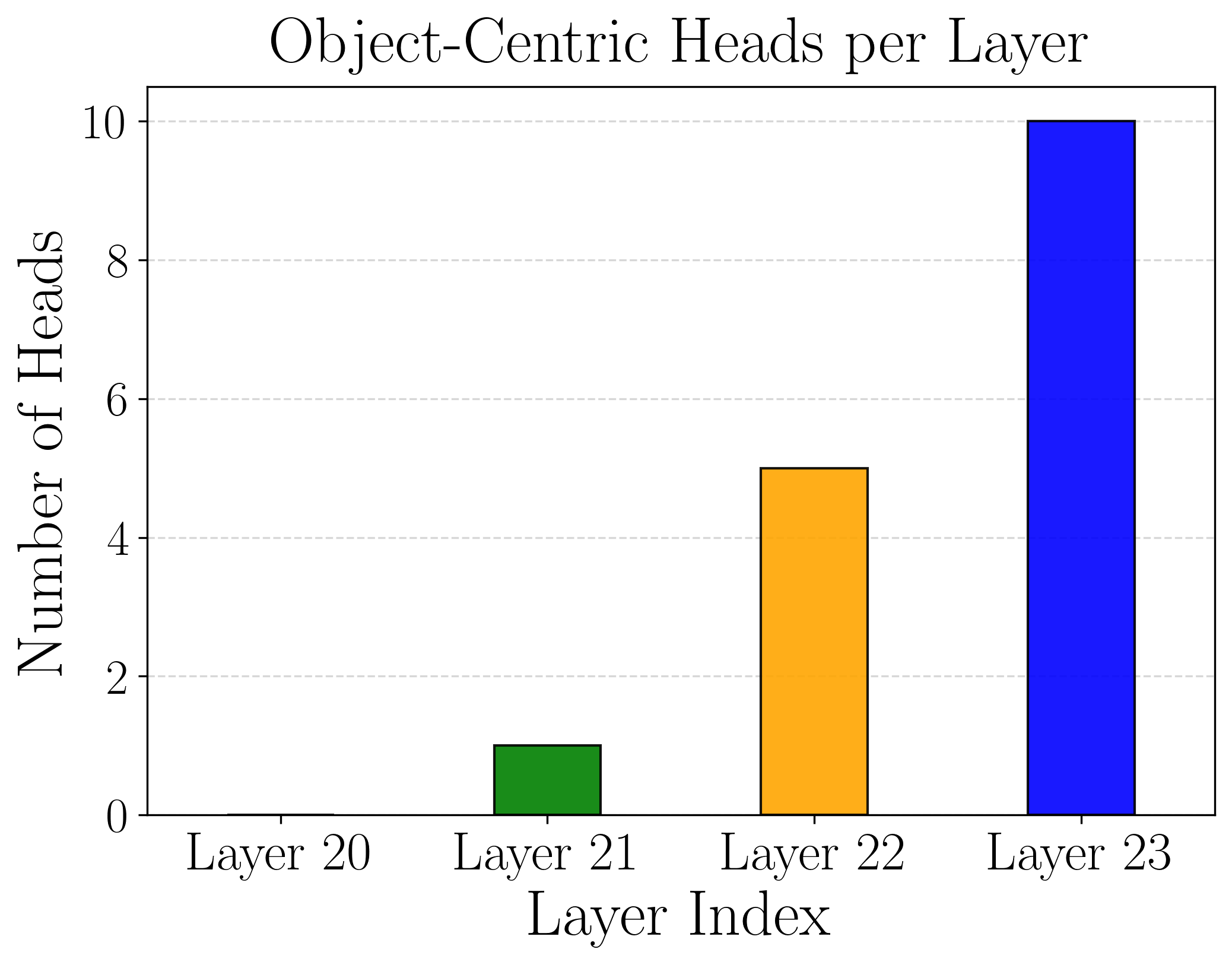}
                \caption*{(c) Highly active heads per layer.}
            \end{minipage}
        \end{tabular}
    }
    \caption{Analysis of the object-centric head distribution in ViT-L, computed over 4,000 images from the COCO dataset. (a) The heatmap shows the frequency of heads (across all 24 layers) belonging to the object-centric cluster. The red boxes highlight that numerous heads in later layers (e.g., Layers 20-23) are consistently selected, demonstrating that object-centric information is a distributed phenomenon and not confined to the final layer. (b) The plot details the final layer's contribution, showing that several heads (circled in black) have low frequency, confirming that some final-layer heads are "noisy" (non-object-centric). (c) The histogram shows the number of strongly active object-centric heads per layer, with Layer 23 containing 10 heads, Layer 22 containing 5 heads, and Layer 21 containing 1 head.}

    \label{supp_fig:analysis_figs}
\end{figure*}

\section{Architectural and Objective Analysis of Object-Centric Heads}
\textbf{Architectural Analysis.} To confirm that object-centric processing is a general property of Vision Transformers, we extended our analysis to ViT-L, a deeper 24-layer, 16-head architecture. The analysis was conducted on the same 4,000-image COCO dataset from the main paper. The results (Figure \ref{supp_fig:analysis_figs}) are consistent with our primary findings:

1. Distributed Object-Centric Heads: We observed that object-centric heads are not isolated in the final layer but are distributed across the network. As with other models, these heads are most heavily concentrated in the later layers (specifically 20-23).

2. Final-Layer Heterogeneity: The final layers are not uniformly object-centric. Layer 23 (Panel b) shows significant variance in head activation frequency (mean = 0.699). Crucially, we again identified several "noisy" heads (circled in black) with substantially lower contributions.

These findings confirm that the observed distribution of specialized and non-specialized heads is a general characteristic across different ViT architectures.

\paragraph{\textbf{Objective Analysis.}} Here, we extend our analysis to Masked Autoencoders (MAE) \cite{mae}, which uses a reconstruction-based self-supervised objective. Unlike DINO's contrastive approach that explicitly encourages semantic similarity through teacher-student distillation, MAE learns representations by reconstructing masked image patches. This fundamental difference in training objectives raises an important question: Does object-centric information still emerge and distribute across the network in reconstruction-based models?

\textbf{Experimental Setup.} We apply our Object-DINO analysis framework to a MAE ViT-B/16 model pre-trained on ImageNet. Following the same protocol as our DINO experiments, we compute the ensemble similarity matrices ($A_ens$) for all attention heads across all layers using 4,000 images from the COCO dataset. We then perform k-means clustering (K=5) to identify object-centric heads.

\begin{figure*}
    \centering
    \begin{tabular}{@{\hskip 0pt}c@{\hskip 2pt}c@{\hskip 2pt}c@{\hskip 2pt}c@{\hskip 2pt}c@{\hskip 0pt}}
    Input Image & Query Similarity ($A_q$)  & Key Similarity  ($A_k$)& Value Similarity ($A_v$) & Ensemble Similarity \\
        \includegraphics[width=0.20\textwidth]{figs/teaser/test_image.jpg} &
        \includegraphics[width=0.20\textwidth]{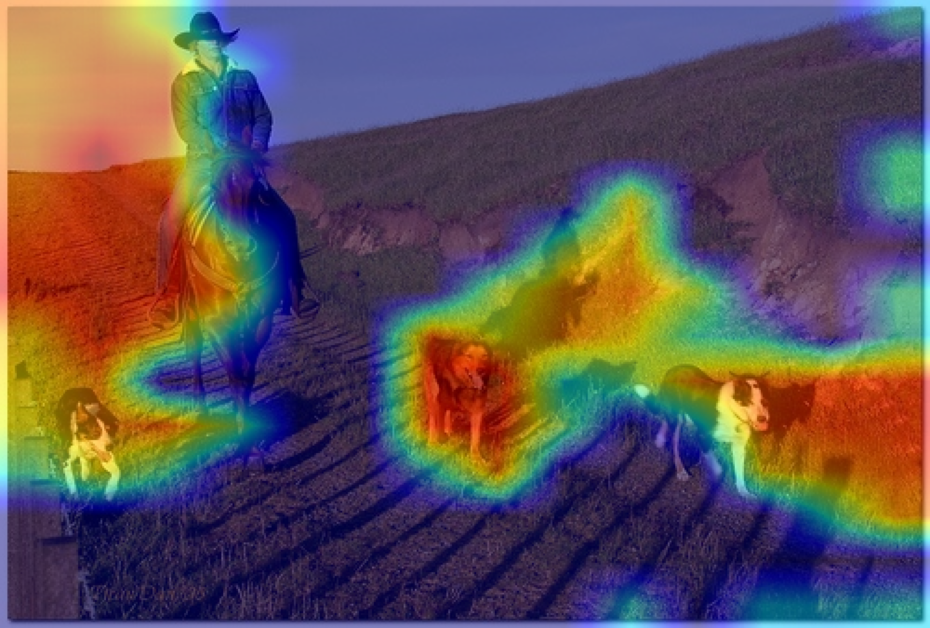} &
        \includegraphics[width=0.20\textwidth]{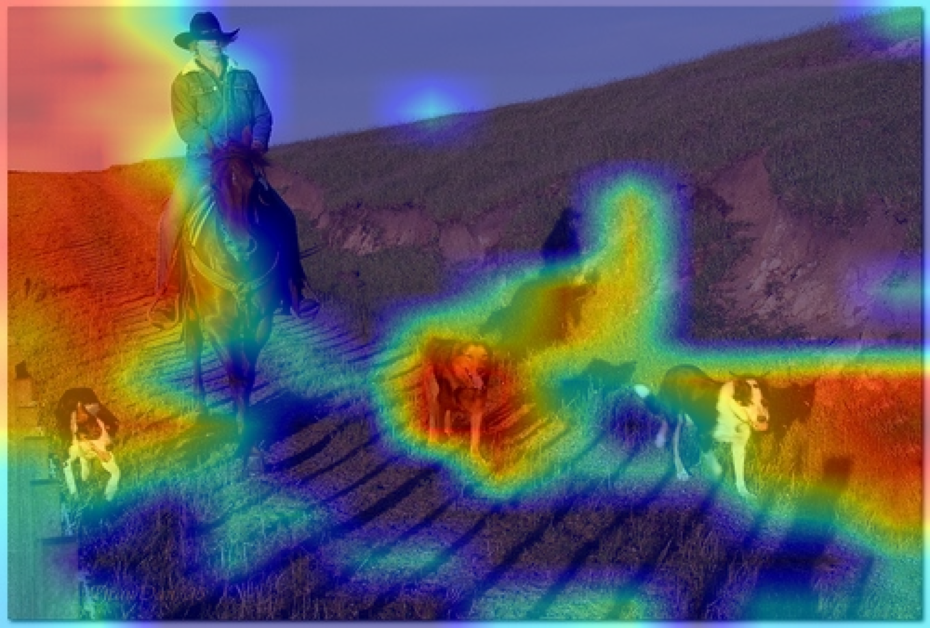} &
        \includegraphics[width=0.20\textwidth]{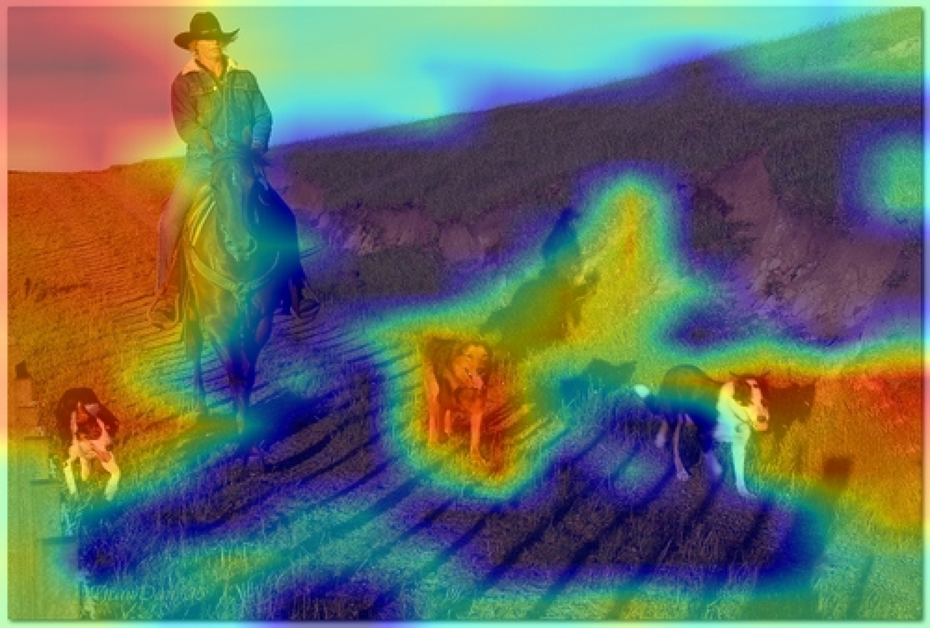} &
        \includegraphics[width=0.20\textwidth]{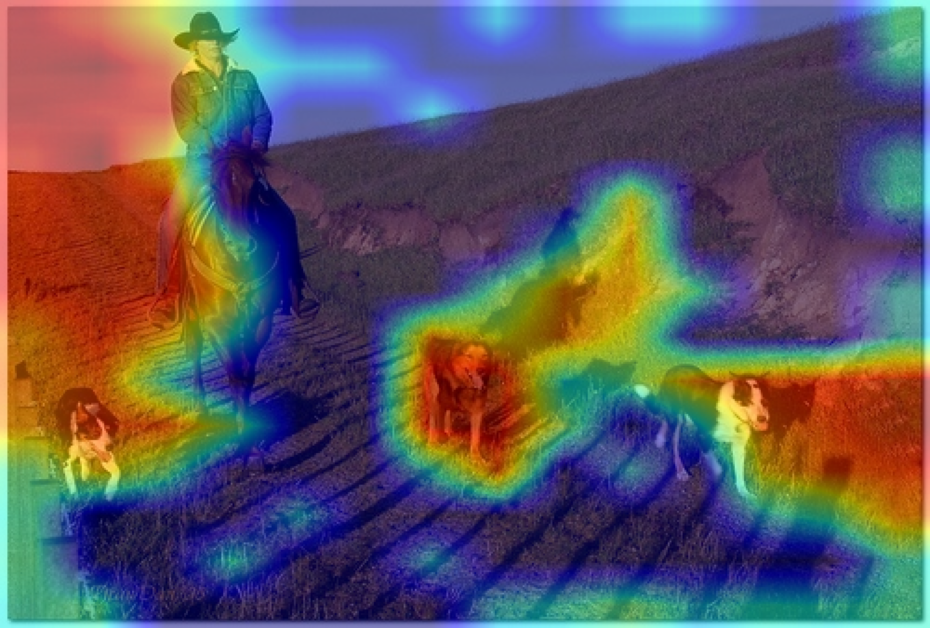} \\
        \includegraphics[width=0.20\textwidth]{figs/teaser/heatmaps_fig4/COCO_val2014_000000139605.jpg} &
        \includegraphics[width=0.20\textwidth]{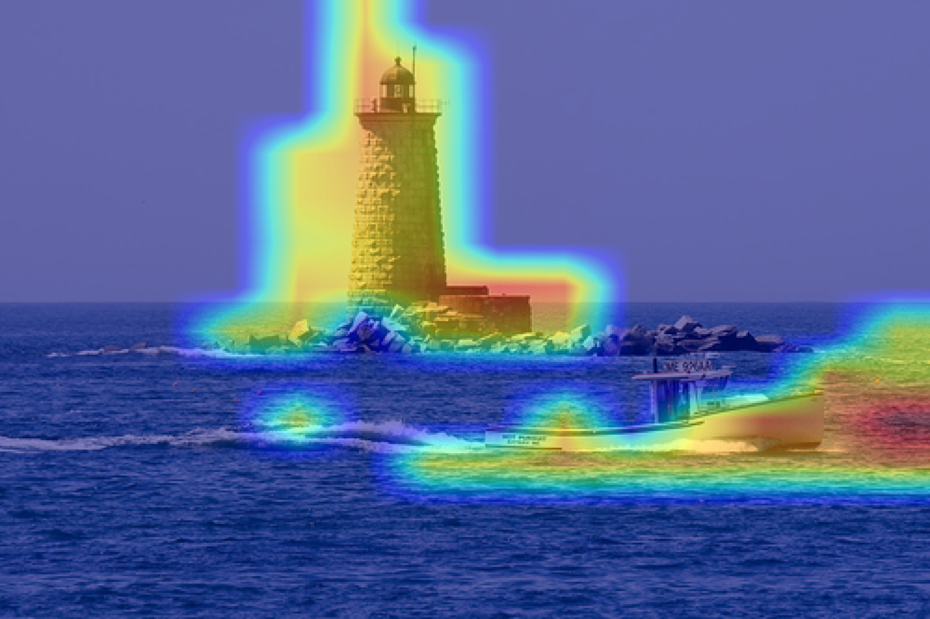} &
        \includegraphics[width=0.20\textwidth]{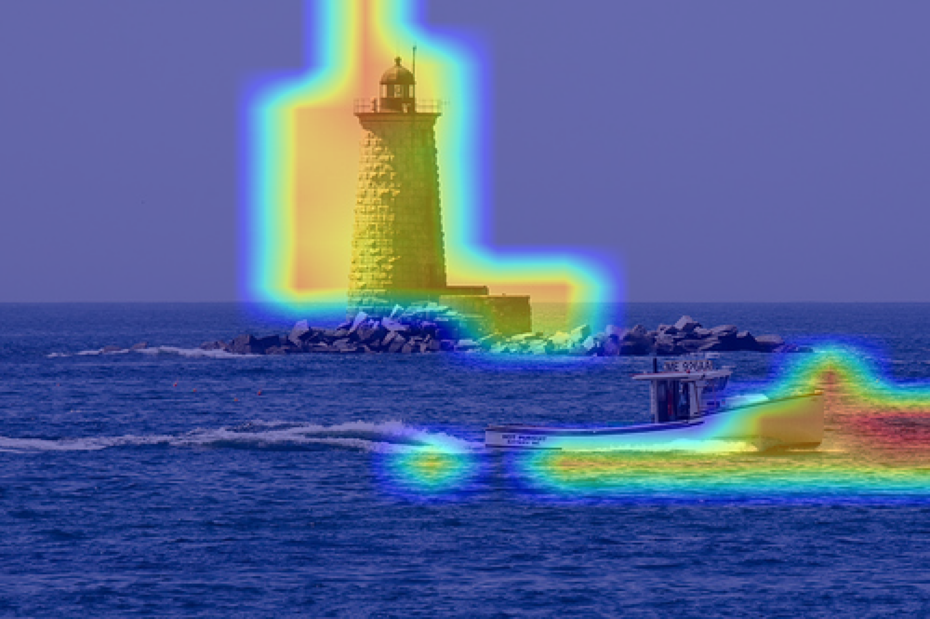} &
        \includegraphics[width=0.20\textwidth]{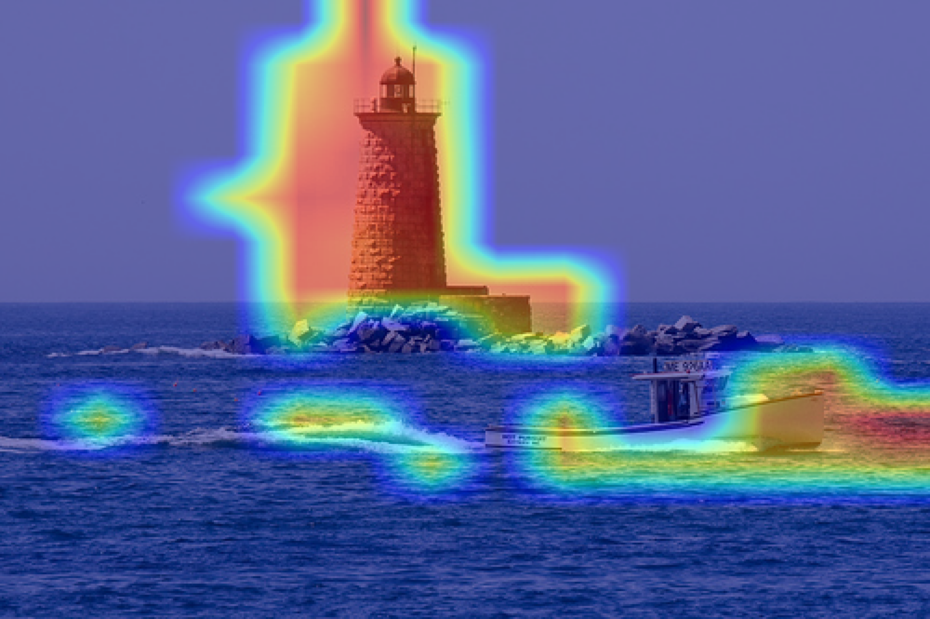} &
        \includegraphics[width=0.20\textwidth]{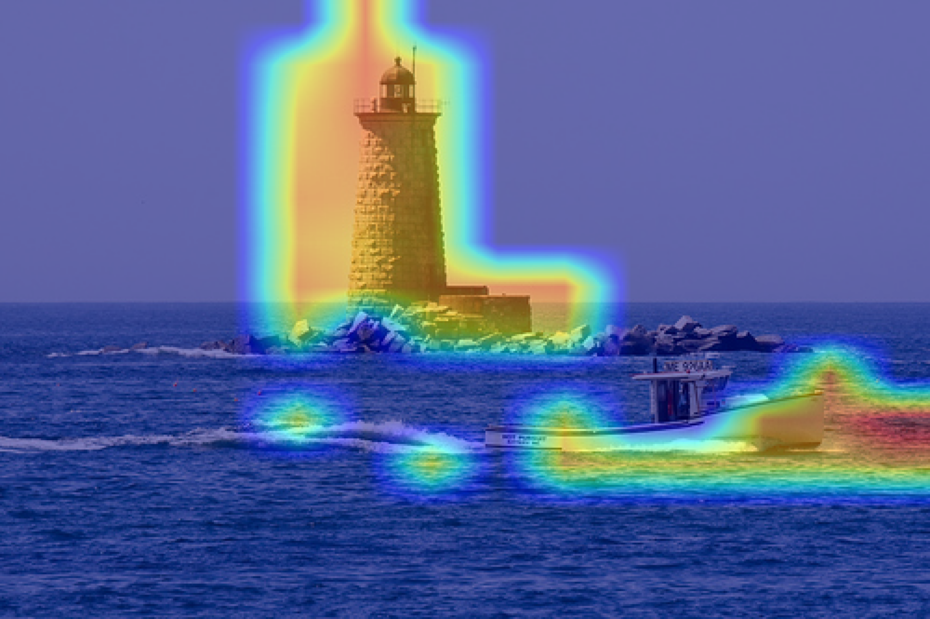} \\
        
    \end{tabular}
    \captionof{figure}{\textbf{Object-centric information in patch-level interactions from MAE \cite{mae}}. We visualize the inter-patch similarity maps ($A_q, A_k, A_v$) computed from the Query, Key, and Value representations of patch tokens in a MAE ViT-B/16 model. For visualization, we invert the similarity maps so objects appear bright. }
    \label{fig:mae_vis}
\end{figure*}

\textbf{Qualitative Analysis.} Figure \ref{fig:mae_vis} shows representative examples of the patch-level similarity maps ($A_q, A_k, A_v$) and their ensemble from a MAE model. We observe that MAE does encode object-centric information in its attention components. The ensemble maps show recognizable foreground-background separation and partial object localization. However, we observe differences compared to DINO. First, the individual component maps ($A_q, A_k, A_v$) are considerably noisier, with more spurious activations in background regions. Second, the ensemble map, while identifying the general object location, does not produce sharp boundaries and often includes substantial background. 

\begin{figure*}[tp]
    \centering
    \resizebox{0.95\textwidth}{!}{%
        \begin{tabular}{ccc}
            \begin{minipage}{0.32\textwidth}
                \centering
                \includegraphics[width=\linewidth,height=0.75\linewidth,keepaspectratio]{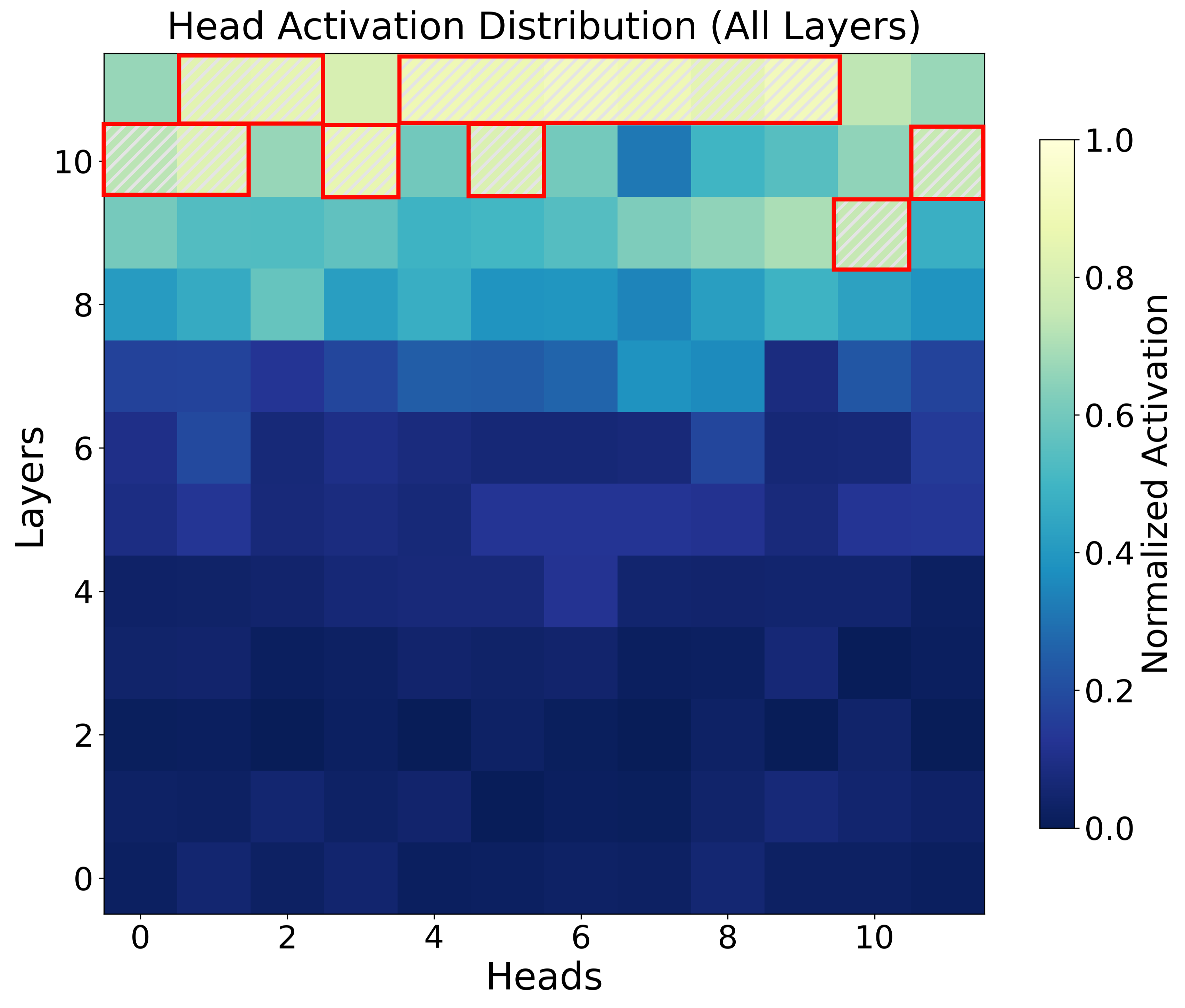}
                \caption*{(a) Head contribution for complete network}
            \end{minipage} &
            \begin{minipage}{0.32\textwidth}
                \centering
                \includegraphics[width=\linewidth,height=0.85\linewidth,keepaspectratio]{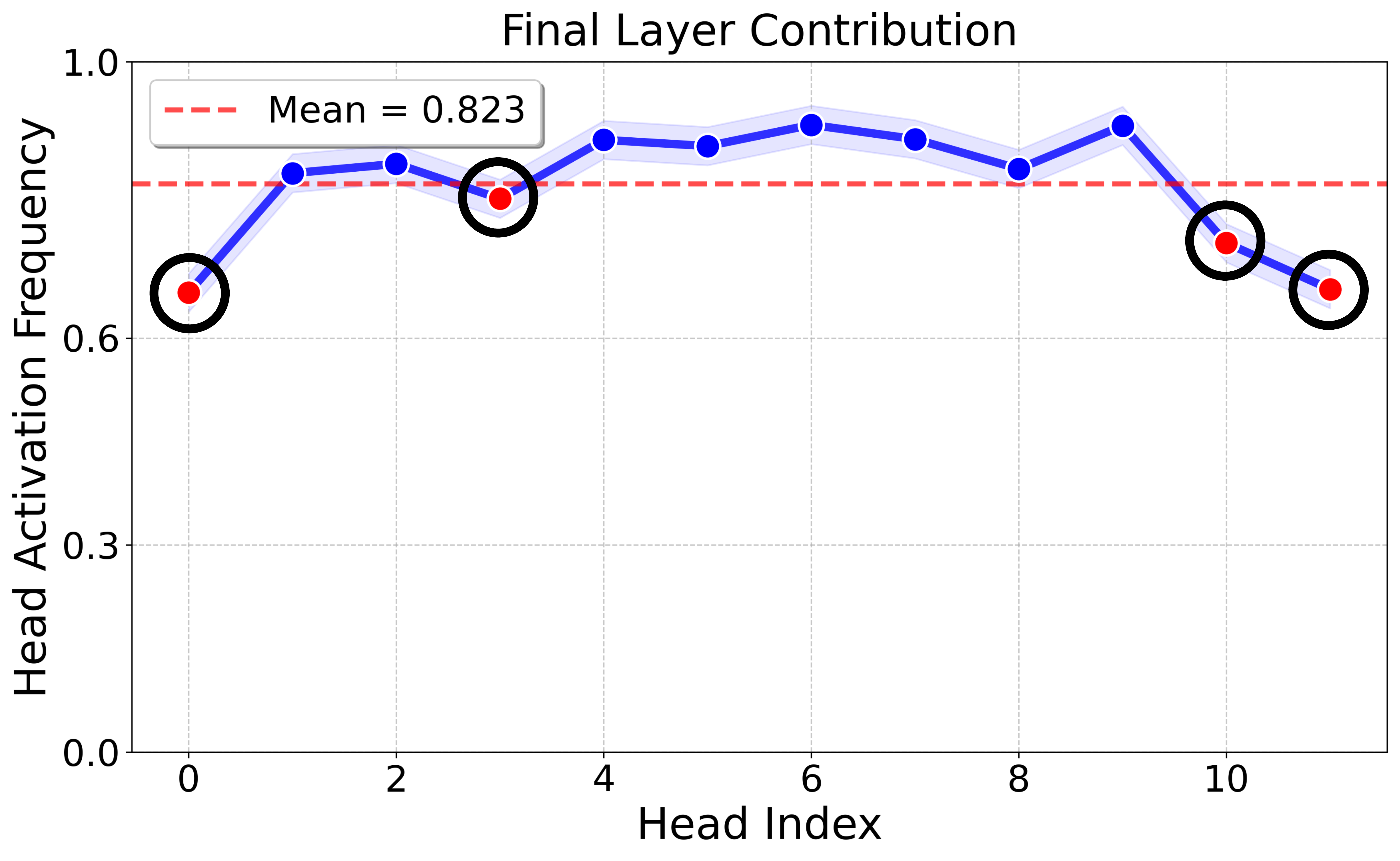}
                \caption*{(b) Heads contribution from final layer.}
            \end{minipage} &
            \begin{minipage}{0.33\textwidth}
                \centering
                \includegraphics[width=0.8\linewidth,height=0.8\linewidth,keepaspectratio]{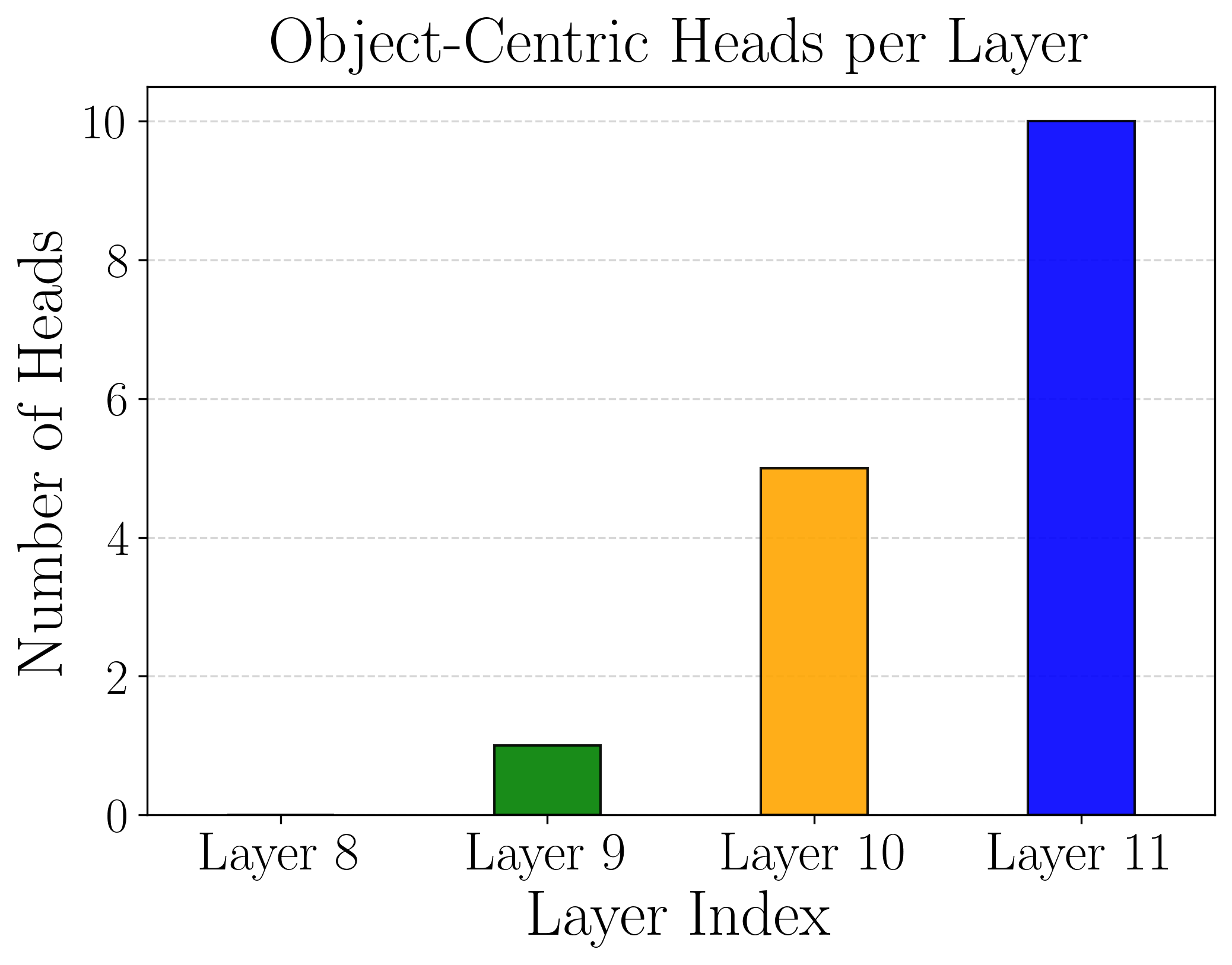}
                \caption*{(c) Highly active heads per layer.}
            \end{minipage}
        \end{tabular}
    }
    \caption{Analysis of the object-centric head distribution in MAE \cite{mae}, computed over 4,000 random images from the COCO dataset. (a) The heatmap shows the frequency of heads (across all 12 layers) belonging to the object-centric cluster identified by our k-means clustering (K=5). (b) The plot shows the final layer's contribution, showing that some heads (circled in black) have low frequency. (c) The histogram shows the number of strongly active object-centric heads per layer.}
    \label{supp_fig:analysis_figs_mae}
\end{figure*}

\textbf{Quantitative Head Distribution Analysis.} Figure \ref{supp_fig:analysis_figs_mae} shows how object-centric heads distribute across MAE's architecture. (a) Shows the frequency heatmap of heads belonging to the object-centric cluster across all 12 layers. Similar to DINO, we observe that object-centric information is distributed throughout the network rather than confined to the final layer. (b) Examines the final layer (Layer 11) specifically and reveals substantial heterogeneity—several heads (circled in black) show low activation frequencies, indicating they are non-object-centric. (c) Quantifies the number of highly active object-centric heads per layer

\section{Evaluation Metrics and Datasets}

In this section, we provide detailed descriptions of the evaluation metrics used throughout our paper. We organize them by task: unsupervised object discovery (CorLoc) and multimodal hallucination mitigation (CHAIR and POPE).

\subsection{CorLoc: Correct Localization}

CorLoc is the standard metric for evaluating unsupervised object discovery methods. It measures the percentage of images where the predicted bounding box correctly localizes at least one ground-truth object instance.

For a dataset $\mathcal{D}$ containing $N$ images, let $B_{\text{pred}}^{(i)}$ denote the predicted bounding box for image $i$, and let $\mathcal{B}_{\text{gt}}^{(i)} = \{B_{\text{gt}}^{(i,1)}, B_{\text{gt}}^{(i,2)}, \ldots, B_{\text{gt}}^{(i,m_i)}\}$ denote the set of $m_i$ ground-truth bounding boxes for that image. The Intersection over Union (IoU) between two boxes is defined as:

\begin{equation}
\text{IoU}(B_1, B_2) = \frac{\text{Area}(B_1 \cap B_2)}{\text{Area}(B_1 \cup B_2)}
\end{equation}

An image is considered correctly localized if the predicted box has IoU $> 0.5$ with at least one ground-truth box:

\begin{equation}
\text{CorLoc} = \frac{1}{N} \sum_{i=1}^{N} \text{Correct}^{(i)} \times 100\%
\end{equation}

\noindent We report CorLoc on three standard benchmarks:
\begin{itemize}
    \item \textbf{PASCAL VOC 2007 \& 2012} \cite{pascal-voc}: These datasets contain images from 20 object categories with bounding box annotations.
    \item \textbf{COCO 20k} \cite{coco20k}: A subset of 20,000 images from the MS-COCO dataset used for evaluating unsupervised discovery methods.
\end{itemize}

\subsection{CHAIR}
CHAIR \cite{chair} quantifies object hallucination in image captioning by measuring how often models mention objects that are not actually present in the input image.

For each image, we extract all objects mentioned in the generated caption and compare them against the ground-truth objects present in the image (from MS-COCO annotations). An object is considered hallucinated if it appears in the caption but not in the ground-truth. CHAIR provides two complementary metrics:

\textbf{CHAIR$_\text{S}$ (Sentence-level):} 

\begin{equation}
\text{CHAIR}_{\text{S}} = \frac{\text{\# captions with hallucinated objects}}{\text{\# all captions}} \times 100\%
\end{equation}

This measures the percentage of captions that contain at least one hallucinated object.

\textbf{CHAIR$_\text{I}$ (Instance-level):} 

\begin{equation}
\text{CHAIR}_{\text{I}} = \frac{\text{\# hallucinated objects}}{\text{\# all mentioned objects}} \times 100\%
\end{equation}

\subsection{POPE}

POPE \cite{POPE} evaluates whether MLLMs can accurately determine object presence through binary yes/no questions. Unlike CHAIR, which evaluates free-form generation, POPE directly tests object recognition.

For each image, POPE generates questions in the format: "Is there a \{object\} in the image?" The questions include both objects that are present (positive samples) and objects that are not present (negative samples). Negative samples are created using three strategies: Random, Popular (frequently co-occurring objects), and Adversarial (objects that typically appear together with present objects).

POPE reports three standard classification metrics: Accuracy, Precision, F1-Score. F1 balances precision (avoiding hallucinations) and recall (detecting actual objects). We evaluate on the standard POPE benchmark constructed from MS-COCO validation images following the standard protocol \cite{POPE}.

\begin{figure*}
    \centering
    \begin{tabular}{@{\hskip 0pt}c@{\hskip 2pt}c@{\hskip 0pt}}
        \includegraphics[width=0.50\textwidth]{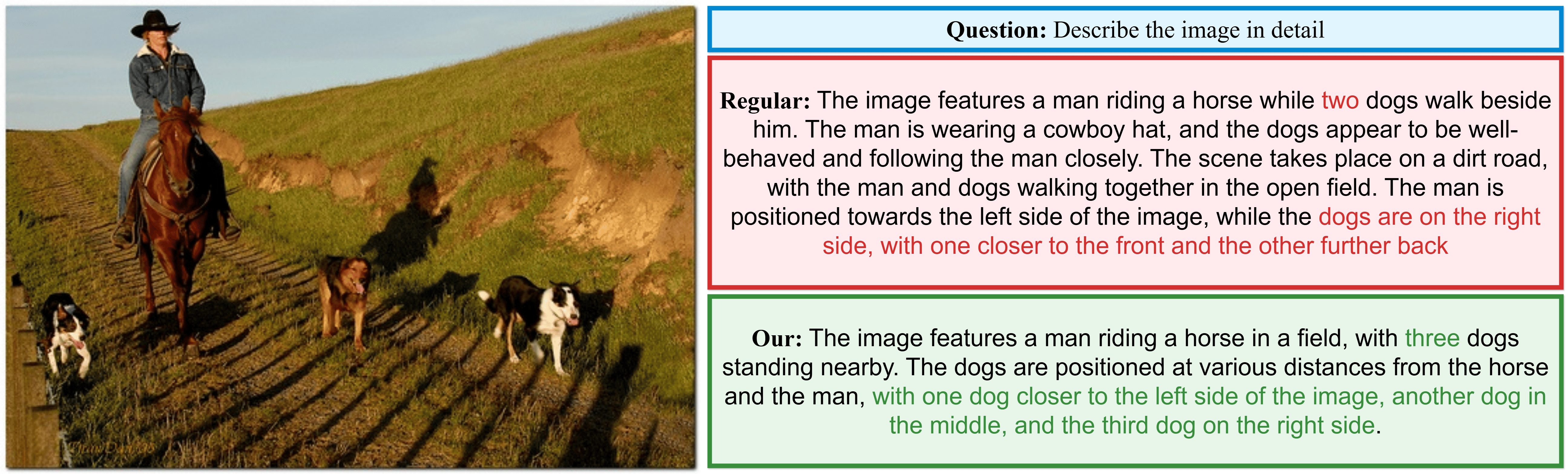} &
        \includegraphics[width=0.485\textwidth]{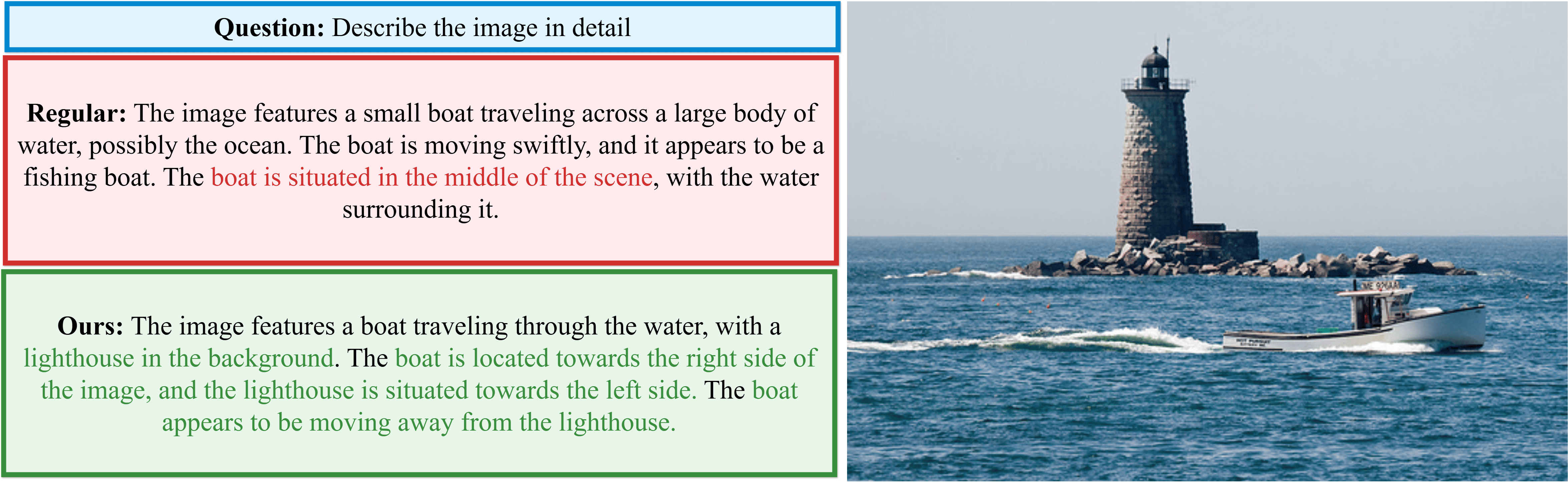} \\

        \includegraphics[width=0.50\textwidth]{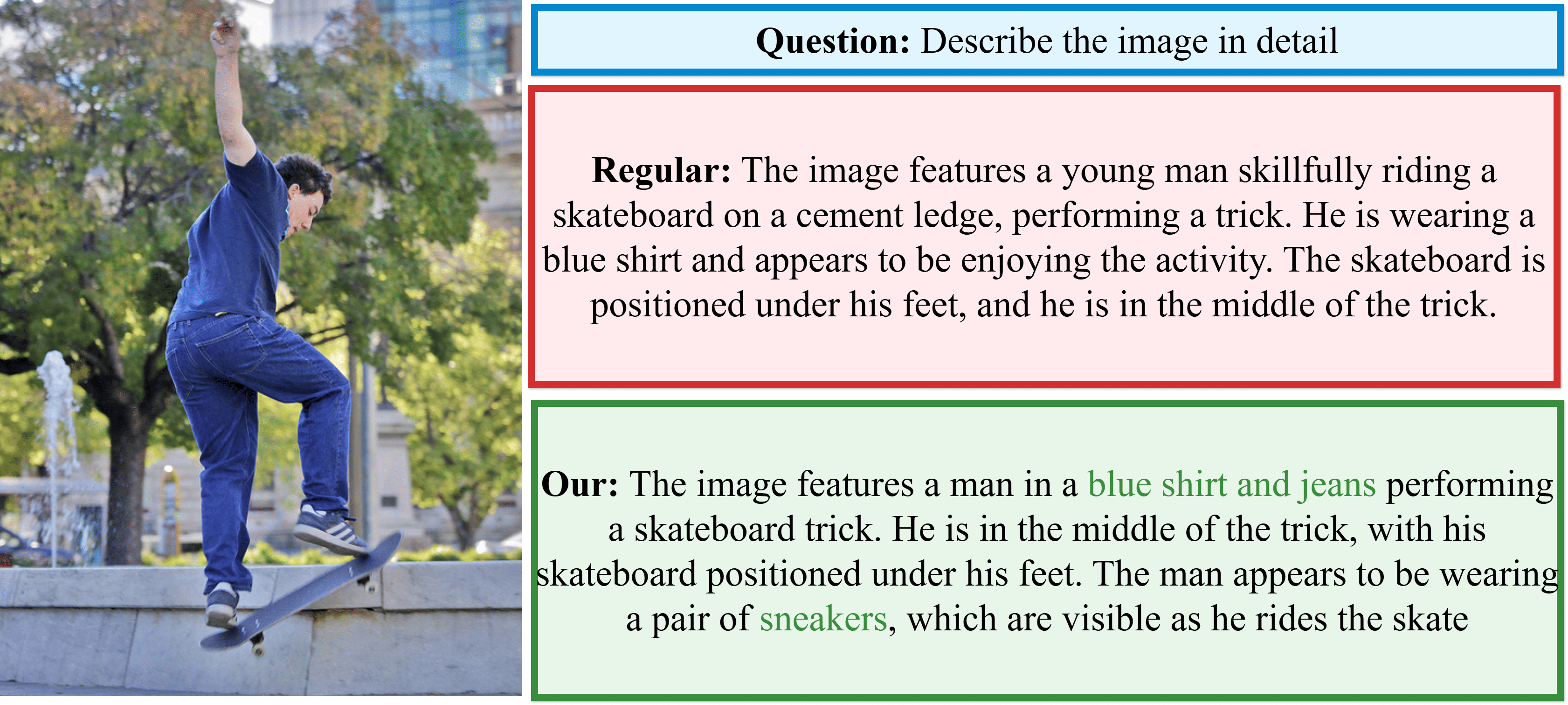} &
        \includegraphics[width=0.475\textwidth]{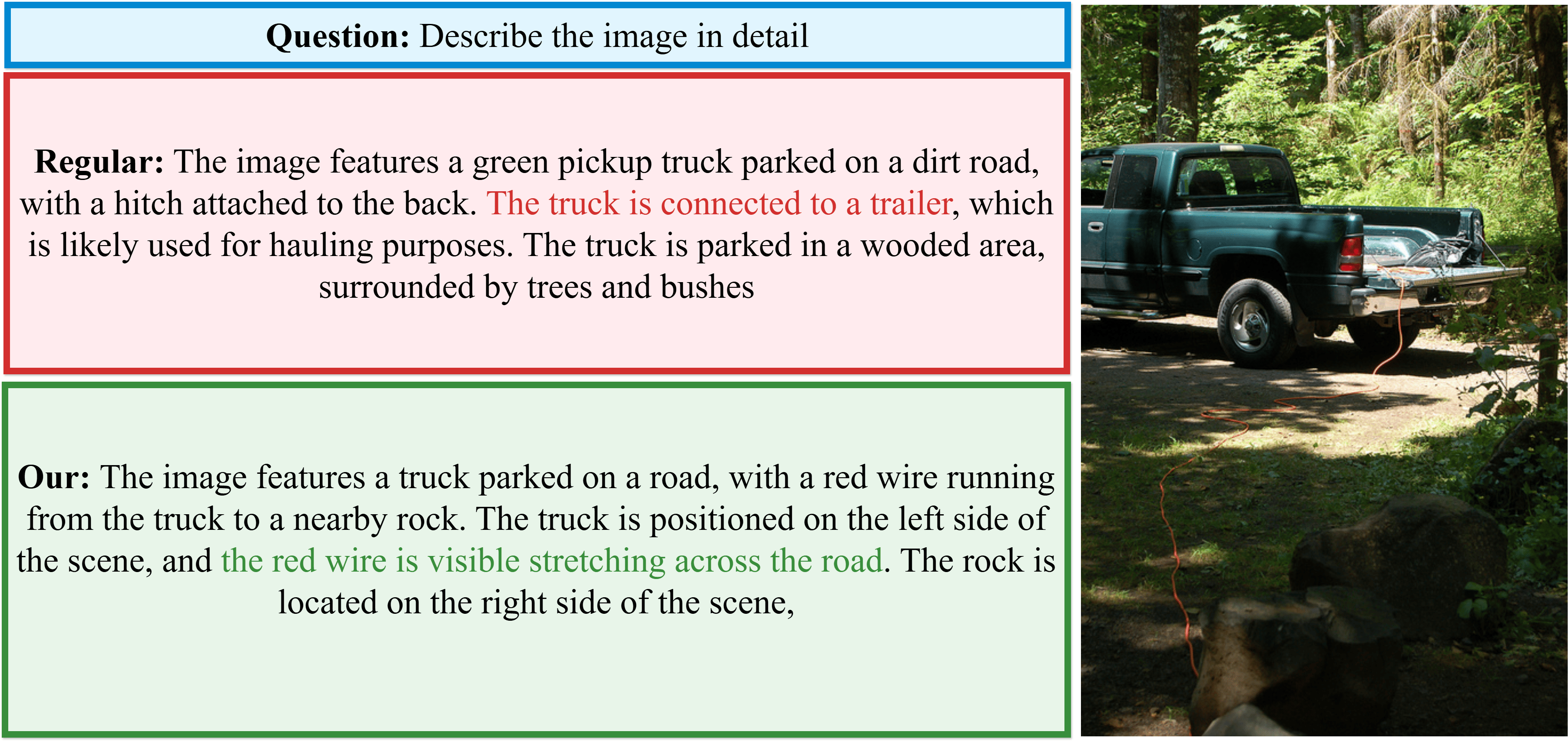}
    \end{tabular}

    \captionof{figure}{Qualitative comparison of captions generated by Regular decoding (Red) and Ours (Green)}
    \label{fig:MLLM_hall_supp}
\end{figure*}


\end{document}